\documentclass[10pt,journal,compsoc]{IEEEtran}
\ifCLASSOPTIONcompsoc
\usepackage[nocompress]{cite}
\else
\usepackage{cite}
\fi
\ifCLASSINFOpdf
\else
\fi

\usepackage{url}
\usepackage{amsmath,amssymb}
\usepackage{color,colortbl}
\definecolor{ballblue}{rgb}{0.13, 0.67, 0.80}
\definecolor{amaranth}{rgb}{0.90, 0.17, 0.31}
\definecolor{olive}{rgb}{0.5, 0.5, 0.0}
\usepackage[colorlinks,citecolor=green]{hyperref}
\usepackage{booktabs} 
\usepackage{graphicx}
\usepackage{dsfont}
\usepackage{arydshln}
\usepackage{multirow}
\usepackage{multicol}
\usepackage{flushend}

\definecolor{Gray}{gray}{0.85}
\newcolumntype{a}{>{\columncolor{Gray}}c}
\definecolor{blush}{rgb}{0.87, 0.36, 0.51}

\begin{document}
	\title{Source Data-absent Unsupervised Domain Adaptation through Hypothesis Transfer and Labeling Transfer}
	\author{Jian Liang,
		Dapeng Hu,
		Yunbo Wang,
		Ran He,~\IEEEmembership{Senior Member,~IEEE,}
		Jiashi Feng,~\IEEEmembership{Member,~IEEE}
		\IEEEcompsocitemizethanks{
			\IEEEcompsocthanksitem J. Liang and R. He are with National Laboratory of Pattern Recognition (NLPR), Institute of Automation, Chinese Academy of Sciences (CASIA), Beijing 100190, China. E-mail: liangjian92@gmail.com, rhe@nlpr.ia.ac.cn.
			\IEEEcompsocthanksitem D. Hu and J. Feng are with the Department of Electrical and Computer Engineering, National University of Singapore (NUS), Singapore 119077, Singapore.
			E-mail: dapeng.hu@u.nus.edu, elefjia@nus.edu.sg.
			\IEEEcompsocthanksitem Y. Wang is with Wangxuan Institute of Computer Technology, Peking University, Beijing 100871, China. Email: wangyunbo09@gmail.com.
		}
		\thanks{Manuscript received 22 Oct. 2020; revised 28 May 2021; accepted 3 Aug. 2021. (Corresponding author: Jian Liang.)}}
	\markboth{IEEE Transactions on Pattern Analysis and Machine Intelligence, doi: no. 10.1109/TPAMI.2021.3103390}%
	{Liang \MakeLowercase{\textit{et al.}}: Source Data-absent Unsupervised Domain Adaptation through Hypothesis Transfer and Labeling Transfer}
	
	\IEEEtitleabstractindextext{%
		\begin{abstract}
			Unsupervised domain adaptation (UDA) aims to transfer knowledge from a related but different well-labeled source domain to a new unlabeled target domain.
			Most existing UDA methods require access to the source data, and thus are not applicable when the data are confidential and not shareable due to privacy concerns.
			This paper aims to tackle a realistic setting with only a classification model available trained over, instead of accessing to, the source data.
			To effectively utilize the source model for adaptation, we propose a novel approach called Source HypOthesis Transfer (SHOT), which learns the feature extraction module for the target domain by fitting the target data features to the frozen source classification module (representing classification hypothesis).
			Specifically, SHOT exploits both information maximization and self-supervised learning for the feature extraction module learning to ensure the target features are implicitly aligned with the features of unseen source data via the same hypothesis.
			Furthermore, we propose a new labeling transfer strategy, which separates the target data into two splits based on the confidence of predictions (labeling information), and then employ semi-supervised learning to improve the accuracy of less-confident predictions in the target domain.
			We denote labeling transfer as SHOT++ if the predictions are obtained by SHOT.
			Extensive experiments on both digit classification and object recognition tasks show that SHOT and SHOT++ achieve results surpassing or comparable to the state-of-the-arts, demonstrating the effectiveness of our approaches for various visual domain adaptation problems.
			Code is available at \url{https://github.com/tim-learn/SHOT-plus}.
		\end{abstract}
		
		}
	\maketitle
	\IEEEdisplaynontitleabstractindextext
	\IEEEpeerreviewmaketitle
	
	\ifCLASSOPTIONcompsoc
	\IEEEraisesectionheading{\section{Introduction}\label{sec:introduction}}
	\else
	\section{Introduction}
	\label{sec:introduction}
	\fi
	\IEEEPARstart{D}{eep} neural networks have achieved remarkable success in a variety of applications across different fields but at the expense of laborious large-scale training data annotation.
	To avoid expensive data labeling, transfer learning \cite{pan2009survey,csurka2017comprehensive,toldo2020unsupervised} is developed to extract the knowledge from one or more source tasks which is then applied to a target task.
	As a typical example, unsupervised domain adaptation (UDA) tackles the problem setting where the learning task in the source domain is sufficiently similar or the same as that in the target domain but labeled data are only available in the source domain during training.
	Recently, UDA methods have been widely applied to boost performance of many tasks like object recognition~\cite{long2015learning,tzeng2017adversarial,saito2018maximum,csurka2017comprehensive}, semantic segmentation~\cite{zhang2017curriculum,hoffman2018cycada,zou2018unsupervised,toldo2020unsupervised}, sentiment classification~\cite{glorot2011domain,peng2018cross}, object detection \cite{chen2018domain,li2020deep}, and person re-identification \cite{deng2018image,wang2019beyond}. 
	Existing UDA methods mainly follow two paradigms to mitigate the gap between source and target domains.
	The first paradigm matches the statistical moments of different feature distributions at different orders to minimize the distributional divergence between domains \cite{sun2016return,zellinger2017central,peng2019moment}.
	For example, the widely used Maximum Mean Discrepancy (MMD) \cite{gretton2007kernel} measure minimizes the distance between weighted sums of all moments from the source and target domains.
	The second paradigm applies adversarial learning~\cite{goodfellow2014generative} with an additional domain classifier to minimize the Proxy $\mathcal{A}$-distance \cite{ben2010theory} between the domains.
	All these methods require to access the source data during learning to adapt the model to the target domain.

	However, nowadays the data often involves user private information, e.g., those on personal phones or from hospital records. 
	Recently, several data protection frameworks have been proclaimed by the European Union (EU) and some governments, among which the General Data Protection Regulation (GDPR), as a typical example, highlights the safety issue of data transfer.
	Accordingly, it may violate the data privacy policy for previous UDA methods to access the source data during learning to adapt.
	To alleviate this issue in the transfer learning field, Hypothesis Transfer Learning (HTL) \cite{kuzborskij2013stability} explore to retain prior knowledge in a form of hypotheses instead of training data inherited from previous tasks.
	Likewise, in this paper, we introduce a \emph{realistic but challenging} source data-absent UDA setting \cite{liang2020we} with only a well-trained source model provided as supervision.
	Different from HTL, here we do not have any labeled data in the target domain for the UDA problem. 
	Our introduced setting also differs from vanilla UDA in that the source model instead of the source data is provided to the target domain for adaptation, making the cross-domain feature-level distribution matching challenging.

	To address this UDA setting, we propose a novel approach called \emph{Source HypOthesis Transfer} (SHOT).
	SHOT follows common deep UDA methods~\cite{ganin2015unsupervised,long2015learning} to utilize an identical network architecture for different domains, consisting of a feature encoding module and a classification module (hypothesis).
	Like~\cite{tzeng2017adversarial,chang2019domain}, SHOT aims to learn a target-specific feature encoding module to generate target data representations that are well aligned with source data representations, but without accessing the source data or the target data labels.
	Intuitively, if the learned target data representations are aligned with the source ones, their classification results from the fixed source classifier (hypothesis) would be highly confident for a certain class, i.e., the classification outputs being close to one-hot vectors.
	We are then motivated to make SHOT adapt the feature encoding module by fine-tuning the source feature encoding module while freezing the source hypothesis, to maximize the mutual information between intermediate feature representations and outputs of the classifier, since information maximization \cite{shi2012information,hu2017learning} can encourage the classifier to assign disparate one-hot outputs to different target feature representations.

	Though target feature representations are encouraged to fit the source hypothesis via information maximization, some semantically wrong matching between target feature representations and source hypothesis may still occur, leading to wrong labels assigned to the target data.
	To alleviate this, we propose to fully exploit the knowledge in the unlabeled target domain by developing two new self-supervised learning schemes.
	First, considering pseudo labels generated by the source classifier for the target data may be noisy, we propose to attain per-class prototype representations for the target domain itself and apply the nearest prototype classifier to obtain more accurate pseudo labels as direct supervision.
	Secondly, inspired by RotNet~\cite{gidaris2018unsupervised} that predicts the absolute rotation of a rotated image, we come up with a relative rotation prediction task to capture the image-specific self-supervision more precisely, i.e. requiring the model to estimate the relative rotation between one original image and its rotated version.
	The two self-supervisions are used to help discard irrelevant semantic information by exploiting the data distribution of the target domain, thus helping learn feature representations that better fit the source hypothesis. 
	In this way, we obtain a target-specific feature encoding module with the source hypothesis as the shared classifier module across domains.

	Since some low-confident predictions generated with the proposed \emph{hypothesis transfer} strategy are possibly inaccurate, we further put forward a \emph{labeling transfer} strategy as a following step, forming a complete two-stage framework called SHOT++ for UDA problems.
	Particularly, we sort the the confidence of the adapted predictions after SHOT and discover an adaptive threshold to automatically divide the whole target data into two splits, i.e., `easy' split with high confidence and `hard' split with low confidence.
	Empirically, these predictions of samples in the `easy' split are reliable.
	Thus, we employ a popular semi-supervised learning algorithm, MixMatch~\cite{berthelot2019mixmatch}, to enable the reliable labeling information from the `easy' split to flow to the `hard' split in the target domain itself.
	It is worth noting that such a labeling transfer strategy can also be applied to the original source model, or even a black-box predictor without knowing the network architecture.

	Experimental results on multiple benchmark datasets clearly demonstrate the proposed SHOT and SHOT++ obtain competitive results with the state-of-the-art, or outperform the state-of-the-art for three different UDA cases, i.e., closed-set~\cite{saenko2010adapting}, partial-set~\cite{cao2018partial}, multi-source~\cite{peng2019moment} problems. 
	The superior results over prior arts in a semi-supervised domain adaptation (SSDA) scenario \cite{saito2019semi} further verify the versatility of the proposed methods.
	The main contributions of this work are summarized as follows.
	\begin{itemize}
		\item We propose a novel framework, Source HypOthesis Transfer (SHOT), for unsupervised domain adaptation with only the source model provided, which is appealing for privacy protection without access to the source data.
		\item SHOT exploits information maximization to learn a target-specific feature encoding module, which provides an implicit perspective on feature alignment.
		\item SHOT further exploits the knowledge in the unlabeled target domain by developing two new kinds of self-supervisions as auxiliary tasks, which further improves the adaptation performance.
		\item We further propose a new labeling transfer strategy by exploiting the confidence of predictions and enforcing the labeling information to flow from `easy' samples to `hard' samples, even allowing adaptation with a \emph{black-box} source model.
		\item Experiments on several benchmarks demonstrate our methods yield results comparable to or outperforming the state-of-the-arts for three unsupervised domain adaptation scenarios and even semi-supervised domain adaptation.
	\end{itemize}
	
	This paper extends our earlier work \cite{liang2020we} in the following aspects.
	Within the hypothesis transfer framework developed in \cite{liang2020we}, we additionally propose one more self-supervision objective to predict the relative rotation, which facilitates learning semantically meaningful representations in the target domain.
	We also propose a new strategy named labeling transfer that only requires the labeling predictions in the target domain.
	Different from \cite{liang2020we}, it even allows adaptation with a \emph{black-box} source model.
	Besides, it can be incorporated with the hypothesis transfer framework, yielding better adaptation results.
	We also expand the experimental evaluation by adding more datasets for each UDA scenario (e.g., PACS~\cite{li2017deeper} for multi-source UDA) and extending our methods further to semi-supervised domain adaptation.
	Finally, we provide a more detailed model analysis to evaluate the proposed approaches, including training stability, parameter sensitivity and qualitative study.

	\section{Related Work}
	\subsection{Unsupervised Domain Adaptation}
	As a typical example of transfer learning \cite{pan2009survey}, unsupervised domain adaptation (UDA) aims to exploit the knowledge in a different but related labeled dataset to help learn a discriminative model for the unlabeled dataset.
	Early UDA methods \cite{zadrozny2004learning,sugiyama2008direct} assume the \emph{covariate shift} with the identical conditional distributions across domains and approximate the target empirical risk by estimating the weight of each source instance and re-weighting the source empirical risk.
	Later, most UDA methods resort to domain-invariant feature transformation \cite{pan2010domain,long2013transfer,liang2018aggregating} or feature space alignment \cite{gopalan2013unsupervised,fernando2013unsupervised,sun2016return} to pursue distribution alignment.
	However, the transferability of these shallow methods is restricted by task-specific structures \cite{long2018transferable}.
	
	Recently, deep neural networks are well explored to learn transferable representations for domain adaptation, in various visual applications like object recognition \cite{gopalan2013unsupervised,csurka2017comprehensive,long2018conditional} and semantic segmentation \cite{hoffman2018cycada,zou2018unsupervised,zhang2019curriculum,toldo2020unsupervised}.
	Based on the relationship of label spaces between source and target domains, UDA scenarios can be categorized into four cases, i.e., closed-set \cite{saenko2010adapting}, partial-set \cite{cao2018partial}, open-set \cite{panareda2017open}, and universal \cite{you2019universal}.
	Among them, the closed-set UDA has received the most research attention, where the source and target label spaces are assumed to be identical.
	Existing deep closed-set UDA methods can be roughly divided into three distinct categories: discrepancy-based, reconstruction-based, and adversarial-based.
	Discrepancy-based approaches minimize a divergence criterion that measures the distance between the source and target data distributions, and some favoring choices include maximum mean discrepancy (MMD) \cite{long2015learning}, high-order central moment discrepancy \cite{zellinger2017central}, contrastive domain discrepancy \cite{kang2019contrastive}, and the Wasserstein metric \cite{courty2017optimal}.
	Reconstruction-based approaches like \cite{ghifary2016deep} utilize reconstruction as an auxiliary task to pursue shared representations for both domains. 
	In addition, some other reconstruction-based methods \cite{bousmalis2016domain,murez2018image} further seek domain-specific reconstruction and cycle consistency to improve the adaptation performance. 
	Inspired by generative adversarial nets \cite{goodfellow2014generative}, adversarial-based approaches determine the distance between different data distributions based on binary classification performance, which in effect corresponds to the Proxy $\mathcal{A}$-distance or $\mathcal{H}$-divergence in the seminal theoretical framework \cite{ben2010theory}.
	Different from marginal distribution alignment using one binary domain classifier in \cite{ganin2015unsupervised}, following methods encourage joint distribution alignment by considering multiple class-wise domain classifiers \cite{pei2018multi} or a semantic multi-output classifier \cite{cicek2019unsupervised,kurmi2019looking} instead of a feature-conditional domain discriminator \cite{long2018conditional}, respectively.
	There are also some other studies investigating batch normalization \cite{cariucci2017autodial,wang2019transferable} and adversarial dropout \cite{saito2018adversarial,lee2019drop} within the network architecture to ensure feature invariance.
	Despite their efficacy, all these methods assume the target user’s access to the source domain, which is not unpractical since the source data may be private and confidential.

	\subsection{Hypothesis Transfer Learning}
	The concept of hypothesis transfer learning (HTL) is first presented by Kuzborskij and Orabona \cite{kuzborskij2013stability}, also with a formal theory.
	Before it, there are a number of transfer learning works \cite{yang2007cross,mansour2009domain,tommasi2010safety} that assume no explicit access to the source data and are empirically successful.
	Generally, HTL is an attractive and efficient framework that assumes access to a given number of source hypotheses and a small set of training samples from the target domain.
	However, like the famous fine-tuning strategy \cite{yosinski2014transferable}, HTL always requires at least a small set of labeled data in the target domain, limiting its applicability to the semi-supervised DA scenario.
	Inspired by HTL, several recent works~\cite{chidlovskii2016domain,liang2019distant} assume absence of the source data and utilize the encoded information as source supervision for the UDA problem.
	In particular, besides target features, \cite{chidlovskii2016domain} requires predictions of target data, and \cite{liang2019distant} requires the mean and variance per-class calculated on source features.
	Both methods adopt a shallow framework like HTL, which are restricted to the original feature structure.
	By contrast, our work fully exploits the end-to-end feature learning module, allowing more flexibility during adaptation.
	There are also two concurrent deep UDA methods~\cite{li2020model,peng2020federated} that attempt not to access the source data during the adaptation process.
	Our approach differs from \cite{li2020model} as we do not need any additional components like a data generator or classifier within the training algorithm;
	\cite{peng2020federated} introduces the first federated DA setting where knowledge is transferred from the decentralized nodes to a new node without any supervision itself and proposes an adversarial-based solution to protect user privacy, but it may fail to tackle the vanilla UDA setting with only one source domain available.
	
	\subsection{Self-supervised Learning}
	Self-supervised learning \cite{jing2020self} offers great feasibility for effectively utilizing unlabeled data by generating and predicting labels from these data.
	The self-supervised task is also known as pretext task.
	A typical workflow\footnote{\url{https://cutt.ly/DfN3rFU}} is to train a model on one or multiple pretext tasks with unlabeled images and then fine-tune the trained model on a variety of practical downstream tasks.
	In addition, pretext tasks can also be jointly trained with supervised learning tasks on labeled data with shared weights like in \cite{carlucci2019domain,zhai2019s4l}.
	Generally, self-supervised methods involve two aspects: pretext task and loss function.
	Some popular image-specific self-supervision tasks include example colorization \cite{zhang2016colorful}, relative position prediction \cite{doersch2015unsupervised}, rotation prediction \cite{gidaris2018unsupervised}, solving jigsaw puzzles \cite{noroozi2016unsupervised};
	on the other hand, contrastive losses \cite{he2020momentum,chen2020simple} and clustering losses \cite{caron2018deep,caron2019unsupervised} focus on the similarity of sample pairs in the representation space, which always provide better performance.
	Some recent studies \cite{xu2019self,sun2019unsupervised,saito2020universal} explore self-supervision for UDA problems and find it beneficial to accomplishing domain alignment.
	By contrast, this paper elegantly designs two different kinds of self-supervisions for UDA problems.
	
	\subsection{Semi-supervised Learning}
	When the domain shift does not exist, the UDA problem naturally becomes a well-studied semi-supervised learning problem.
	Many ideas originally proposed for semi-supervised learning thus can also be employed to achieve or compensate domain alignment within UDA methods.
	Pseudo-labeling \cite{lee2013pseudo} is a simple heuristic widely used in practice, which produces `pseudo-labels' for unlabeled data using the prediction function itself during the course of training.
	Among UDA methods, \cite{zhang2018collaborative} directly incorporates pseudo-labeling as a regularization term, and \cite{long2018conditional} leverages pseudo labels in the adaptation module to achieve multi-modal distribution alignment.
	Entropy minimization \cite{grandvalet2005semi} is a popular strategy that encourages the network to make `confident' (low-entropy) predictions for all unlabeled data, which has been exploited in many previous UDA methods \cite{long2018transferable,xu2019larger}.
	Other favored semi-supervised techniques like tri-training and virtual adversarial training have been used in frameworks \cite{saito2017asymmetric,shu2018dirt}, respectively.
	Recently, \cite{rukhovich2019mixmatch} directly employs MixMatch \cite{berthelot2019mixmatch} and obtains promising results in the VisDA-2019 challenge.
	Different from prior works that treat the whole target domain as an unlabeled dataset, we focus on intra-domain semi-supervised learning where the labeled dataset consists of confident target data samples and the unlabeled dataset consists of remaining samples.
	
	\section{Method}
	We aim to address the UDA problem with only a pre-trained source model, not requiring to access the source data. 
	In particular, we consider the $K$-way visual classification task.
	For a vanilla UDA task, we are given $n_s$ labeled samples $\{x_s^{i},y_s^{i}\}_{i=1}^{n_s}$ from the source domain $\mathcal{D}_s$ where $x_s^{i} \in \mathcal{X}_s$, $y_s^{i}\in \mathcal{Y}_s$, and also $n_t$ unlabeled samples $\{x_t^{i}\}_{i=1}^{n_t}$ from the target domain $\mathcal{D}_t$ where $x_t^{i} \in \mathcal{X}_t$.
	The goal of UDA is to predict the labels $\{y_t^{i}\}_{i=1}^{n_t}$ in the target domain, where $y_t^{i} \in \mathcal{Y}_t$, and the source task $\mathcal{X}_s \to \mathcal{Y}_s$ is assumed to be the same with the target task $\mathcal{X}_t \to \mathcal{Y}_t$.
	In this work, we aim to learn a target function $f_t: \mathcal{X}_t \to \mathcal{Y}_t$ and infer $\{y_t^{i}\}_{i=1}^{n_t}$, with only $\{x_t^{i}\}_{i=1}^{n_t}$ and the source function $f_s: \mathcal{X}_s \to \mathcal{Y}_s$ available.
	
	We address the above source data-absent UDA problem through the following steps.
	First, we train the classification model, consisting of a feature encoding module and a hypothesis module, from the source data and then transfer the source model to the target domain without accessing the source data.
	Then, we present a novel framework, Source HypOthesis Transfer (SHOT), to learn the target-specific feature encoding module using self-supervised learning and semi-supervised learning, with the source hypothesis fixed.
	Finally, using the predictions for the target domain, we further employ a semi-supervised learning algorithm to enforce labeling information propagation from confidently labeled target samples to the remaining target samples with low confidences. 
	Applying such a labeling transfer strategy to SHOT yields SHOT++.
	Likewise, applying the labeling transfer strategy to `Source-model-only' yields `Source-model-only++', which can even deal with a black-box source model.
	In the following, we elaborate on each step in details. 
	
	\subsection{Source Model Generation}
	We consider learning a deep source classification model $f_s: \mathcal{X}_s \to \mathcal{Y}_s$ by minimizing the following cross-entropy loss,
	\begin{equation}
		\begin{aligned}
			\mathcal{L}_{src}(f_s;\mathcal{X}_s,\mathcal{Y}_s) =& \\
			\mathbb{E}_{(x_s,y_s)\in \mathcal{X}_s \times \mathcal{Y}_s}& \sum\nolimits_{k=1}^{K}- q_k \log \delta_k(f_s(x_s)),
		\end{aligned}
		\label{eq:cross}
	\end{equation}
	where $\delta_k(a)=\frac{\exp(a_k)}{\sum_i \exp(a_i)}$ denotes the $k$-th element in the soft-max output of a $K$-dimensional vector $a$, and $q$ denotes a one-hot encoding of $y_s$ where $q_k$ is `1' for the correct class and `0' for the rest.
	To further lift the discriminability of the source model and facilitate the following target data alignment, we adopt the label smoothing technique for model training as it encourages learned feature representations to form tight and evenly separated clusters \cite{muller2019does}, which is useful for adaptation. 
	Therefore, the source objective function is changed to
	\begin{equation}
		\begin{aligned}
			\mathcal{L}_{src}^{ls}(f_s;\mathcal{X}_s,\mathcal{Y}_s) =& \\
			\mathbb{E}_{(x_s,y_s)\in \mathcal{X}_s \times \mathcal{Y}_s}& \sum\nolimits_{k=1}^{K} - q_k^{ls} \log \delta_k(f_s(x_s)),
		\end{aligned}
		\label{eq:ls}
	\end{equation} 
	where $q^{ls}_k=(1-\alpha)q_k + \alpha/K$ is the smoothed label and $\alpha$ is the smoothing parameter which is empirically set to 0.1.
	
	\subsection{Hypothesis Transfer with Information Maximization}
\begin{figure}[t]
		\centering
		\includegraphics[width=0.45\textwidth]{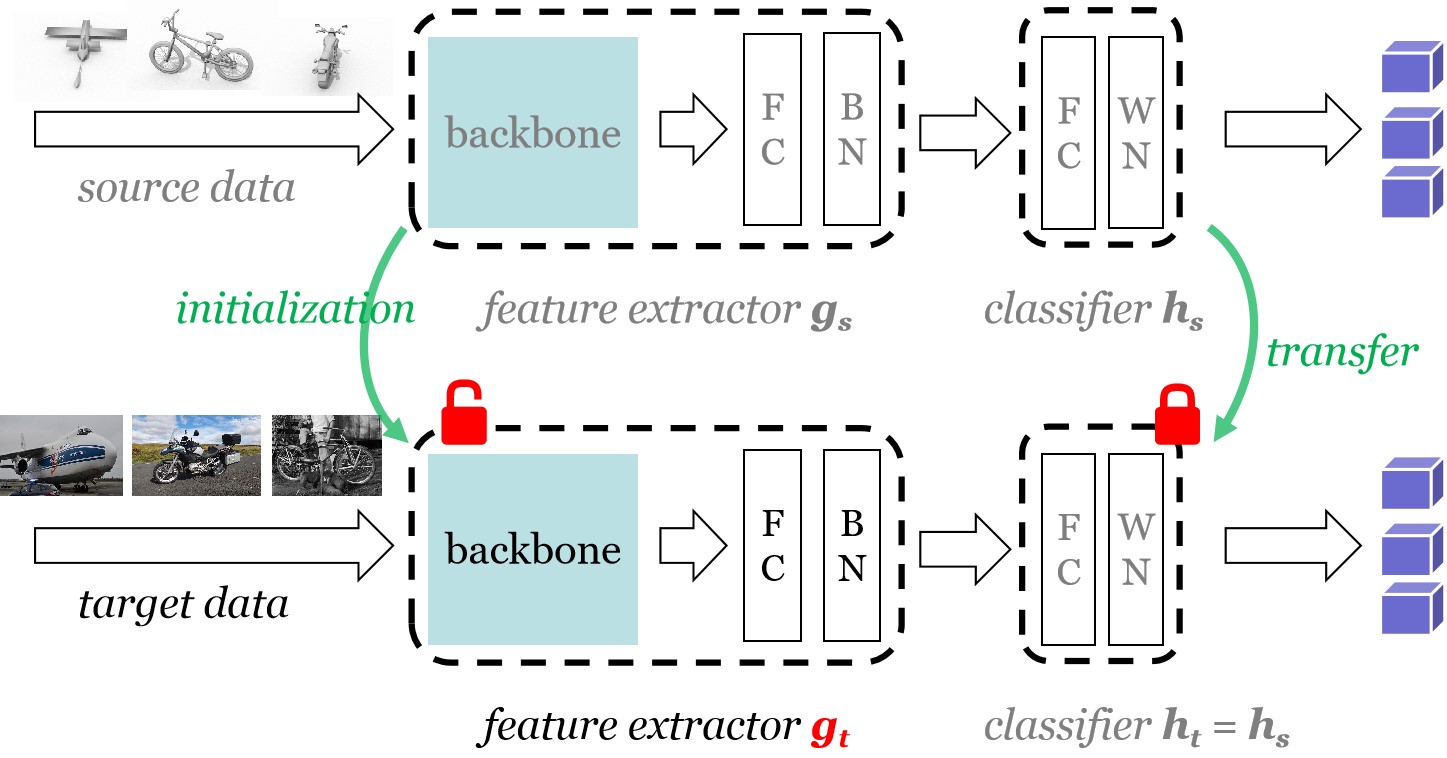}
		\caption{The pipeline of hypothesis transfer with information maximization. The source model consists of a feature encoding module and a classifier module (hypothesis). SHOT keeps the hypothesis frozen and utilizes the feature encoding module as initialization for target domain learning.} 
		\label{fig:framework}
	\end{figure}

	As shown in Fig.~\ref{fig:framework}, the source model parameterized by a deep neural network consists of two modules: the feature encoding module $g_s: \mathcal{X}_s\to \mathbb{R}^{d}$ and the classifier module $h_s: \mathbb{R}^{d}\to \mathbb{R}^{K}$, i.e., $f_s(x) = h_s\left(g_s(x)\right)$, where $d$ is the dimension of the input feature. 
	Most previous UDA methods align different domains by matching the data distributions in the feature space $\mathbb{R}^{d}$ using MMD \cite{long2015learning} or domain adversarial alignment \cite{ganin2015unsupervised}.
	However, both strategies assume the source and target domains share the same feature encoder and need to access the source data during adaptation.
	This is not applicable in the tackled UDA setting here. 
	By contrast, Adversarial Discriminative Domain Adaptation (ADDA) \cite{tzeng2017adversarial} relaxes the parameter-sharing constraint and is a new adversarial framework, which learns different mapping functions for the two domains. 
	Also, Decision-boundary Iterative Refinement Training with a Teacher (DIRT-T) \cite{shu2018dirt} first trains a parameter-sharing UDA framework as initialization and then fine-tunes the whole network by minimizing the cluster assumption violation via entropy minimization and virtual adversarial training.
	Both methods suggest that learning a domain-specific feature encoding module for $\mathcal{D}_t$ is practicable and even works better than the parameter-sharing mechanism, which has also been proven effective in Domain-Specific Batch Normalization (DSBN) \cite{chang2019domain}.

	We therefore develop a new framework termed Source HypOthesis Transfer (SHOT) by learning the domain-specific feature encoding module for the target data while fixing the source classifier module (hypothesis), as the source hypothesis encodes the distribution information of the unseen source data.
	Namely, SHOT utilizes the \emph{same} classifier module $h_t=h_s$ for different domain-specific feature encoding modules. 
	It aims to learn the optimal target feature encoding module $g_t: \mathcal{X}_t\to \mathbb{R}^{d}$ such that the output target features can fit the source feature distribution well and can be accurately classified by the source hypothesis directly.
	Note that SHOT merely utilizes the source data for just once to generate the source hypothesis, and does not need to access the source data any more, unlike prior methods (e.g., ADDA, DIRT-T, and DSBN).

	Essentially, we expect to learn the optimal target feature encoder $g_t$ so that the target data distribution $p\left(g_t(x_t)\right)$ matches the source data distribution $p\left(g_s(x_s)\right)$ well.
	However, feature-level alignment does not work at all since it is impossible to estimate the distribution of $p\left(g_s(x_s)\right)$ without access to the source data.
	We view the challenging problem from another perspective: \emph{if there is no domain gap, what kind of outputs should be generated over the unlabeled target data?}
	We argue the ideal outputs of target features should be similar to those of source features with the classifier shared for both domains.
	Since we train the source feature encoding module $g_s$ and classifier module $h_s$ via a supervised learning loss, the output of each source feature is fairly similar to one of the one-hot encodings.
	Therefore, we expect that the output of each target feature through $h_t=h_s$ is also similar to one of the one-hot encodings.
	Such an output alignment requirement is a necessary condition for feature alignment.

	For this purpose, we adopt the information maximization (IM) loss \cite{krause2010discriminative,shi2012information,hu2017learning}
	to make the classification outputs of target features individually certain and globally diverse.
	In practice, we minimize the following $\mathcal{L}_{ent}$ and $\mathcal{L}_{div}$ that together constitute the IM loss ($\beta=1$): 
	\begin{equation}
		\begin{aligned}
			\mathcal{L}_{im}(f_t;\mathcal{X}_t) & = \mathcal{L}_{ent}(f_t;\mathcal{X}_t) + \beta \mathcal{L}_{div}(f_t;\mathcal{X}_t)\\
			\mathcal{L}_{ent}(f_t;\mathcal{X}_t) &= -\mathbb{E}_{x\in\mathcal{X}_t} \sum\nolimits_{k=1}^{K} \delta_k(f_t(x)) \log \delta_k(f_t(x)),\\
			\mathcal{L}_{div}(f_t;\mathcal{X}_t) &= \sum\nolimits_{k=1}^{K} \hat{p}_k \log \hat{p}_k \\
			&= D_{KL}(\hat{p}, \frac{1}{K}\mathbf{1}_K) - \log K,
		\end{aligned}
		\label{eq:im}
	\end{equation}
	where $f_t(x)=h_t(g_t(x))$ is the $K$-dimensional output of each target sample, $\mathbf{1}_K$ is a $K$-dimensional vector with all ones, and $\hat{p} = \mathbb{E}_{x\in\mathcal{X}_t} [\delta(f_t(x))]$ is the mean output embedding of the whole target domain.
	The IM loss would work better than conditional entropy minimization \cite{grandvalet2005semi} widely used in prior UDA methods \cite{vu2019advent,saito2019semi} since IM can circumvent the trivial solution where all unlabeled data have the same one-hot encoding via the fair diversity-promoting objective $\mathcal{L}_{div}$. 
	For convenience, we denote SHOT with the information maximization loss as SHOT-IM.
	
	\begin{figure}[t]
		\centering
		\footnotesize
		\setlength\tabcolsep{0mm}
		\renewcommand\arraystretch{0.1}
		\begin{tabular}{cc}
			\includegraphics[width=0.49\linewidth,trim={4.4cm 10.0cm 4.4cm 10.0cm}, clip]{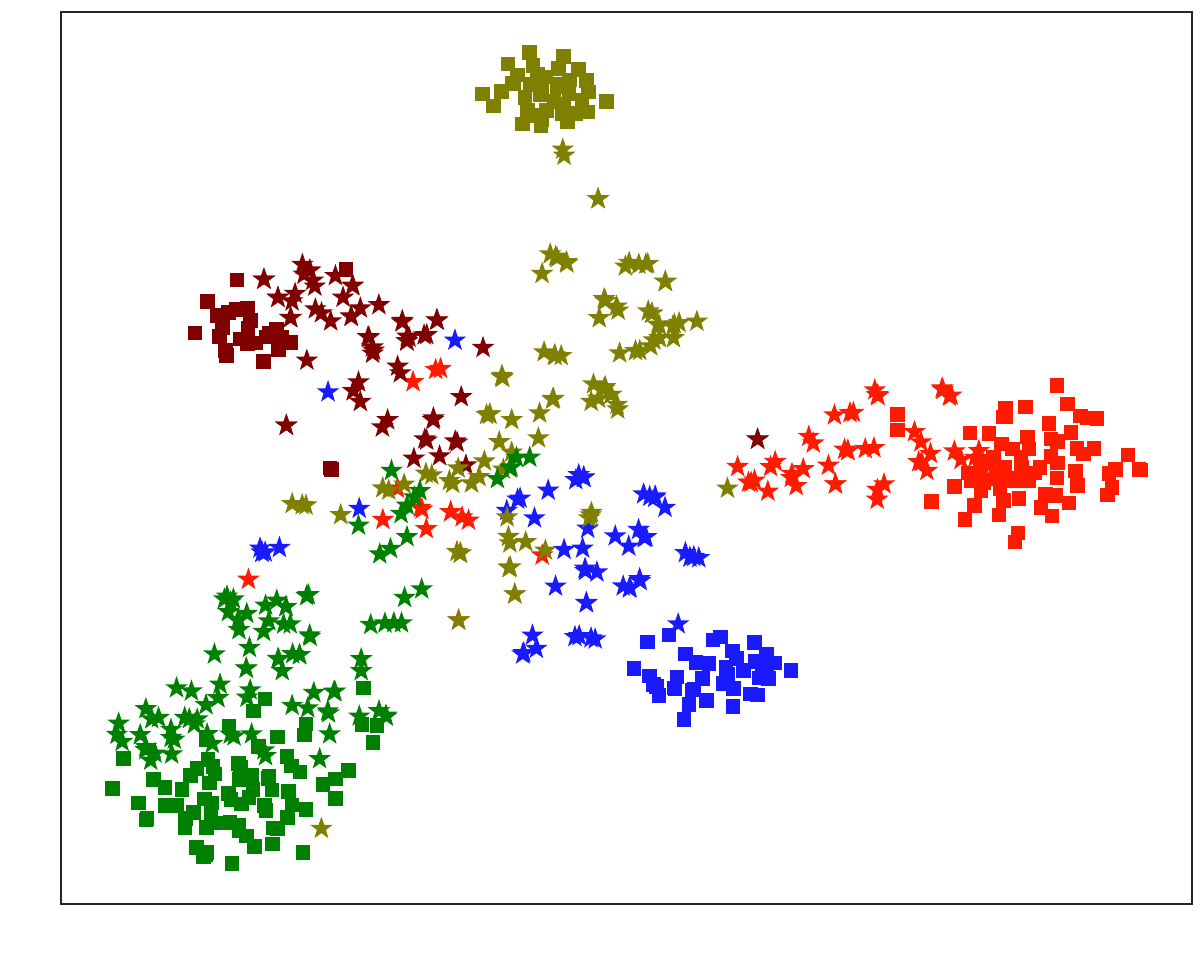} & 
			\includegraphics[width=0.49\linewidth,trim={4.4cm 10.0cm 4.4cm 10.0cm}, clip]{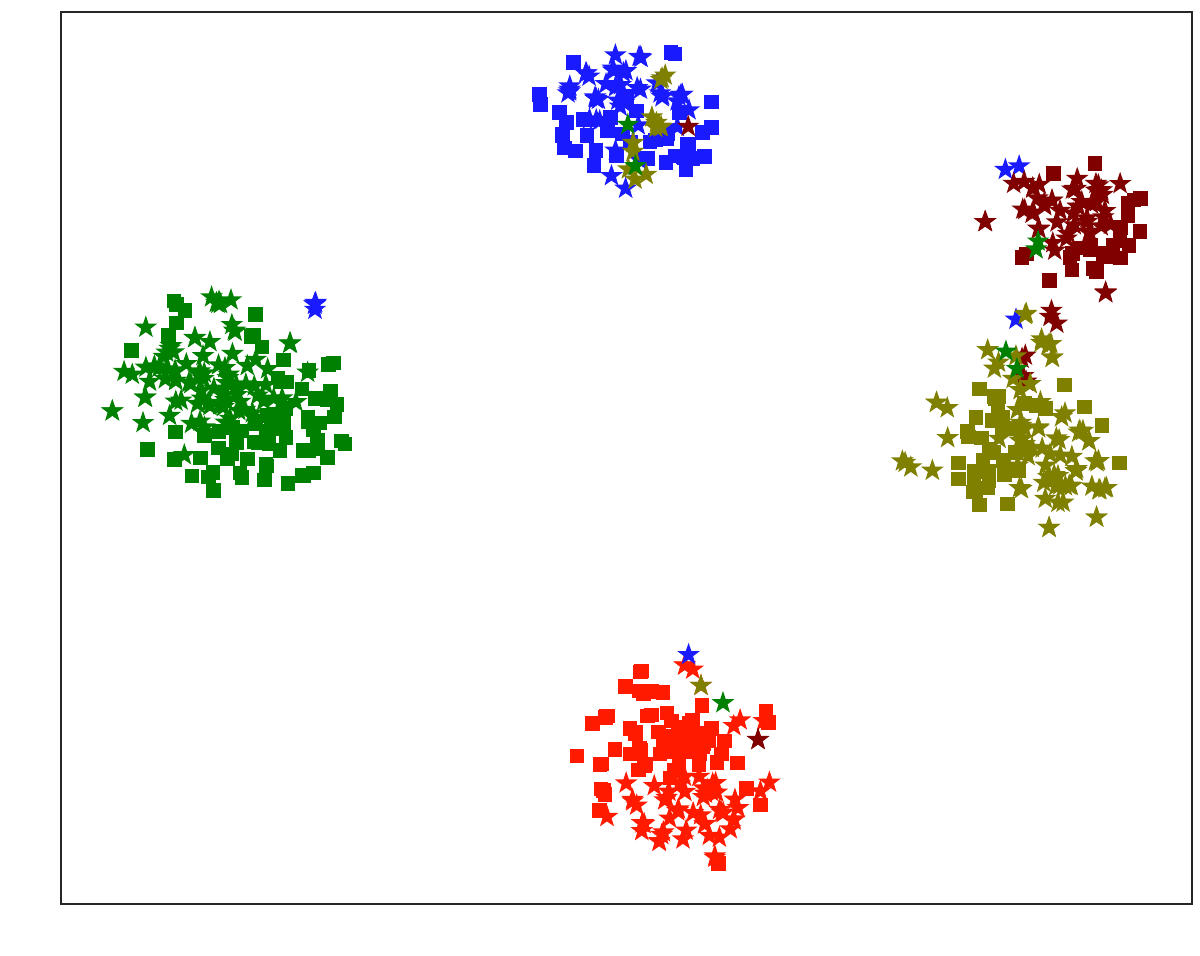} \\
			~\\
			(a) Source model only & (b) SHOT-IM 
		\end{tabular}
		\caption{The t-SNE visualizations for a 5-way classification task. Solid `$\circ$' denotes unseen source data and `$\star$' denotes target data. Different colors represent different classes. Best viewed in colors.}
		\label{fig:tsne}
	\end{figure}

	\subsection{Hypothesis Transfer with Self-supervised Learning}
	Fig.~\ref{fig:tsne} shows the t-SNE visualizations of features for a 5-way classification task learned by SHOT-IM and the `source model only' method.
	Intuitively, the target feature representations are distributed in a mess for the `source model only' method in Fig.~\ref{fig:tsne}(a), and using the IM loss indeed helps align the target data with the unseen source data well.
	However, the target data may be matched to the wrong source hypothesis to some extent in Fig.~\ref{fig:tsne}(b).

	We argue that the harmful effects result from the inaccurate original network outputs. 
	For instance, a target sample from the second class with the normalized network output [0.4, 0.3, 0.1, 0.1, 0.1] may be forced to have an expected output [1.0, 0.0, 0.0, 0.0, 0.0].
	Motivated by \cite{caron2018deep,wallace2020extending}, self-supervised learning helps focus on semantically meaningful features, which is in line with domain invariant learning. 
	Therefore, we try to learn structure-aware and semantic representations in the unlabeled target domain to alleviate such effects.
	Specifically, we develop two new self-supervision objectives to be jointly trained with the main unsupervised task in Eq.~(\ref{eq:im}) in a similar manner to prior methods \cite{zhai2019s4l,sun2019unsupervised}.
	We first exploit self-supervision from the perspective of the loss function and design a novel self-supervised pseudo-labeling strategy.
	Different from pseudo-labeling \cite{lee2013pseudo} where pseudo labels conventionally generated by source hypotheses are still noisy due to domain shift, our self-supervised version considers the structure of the target domain (i.e. the target-specific prototypes) and is able to provide accurate pseudo labels.
	The detailed learning procedure is provided in the following.
	
	\begin{figure}[t]
		\centering
		\includegraphics[width=0.45\textwidth]{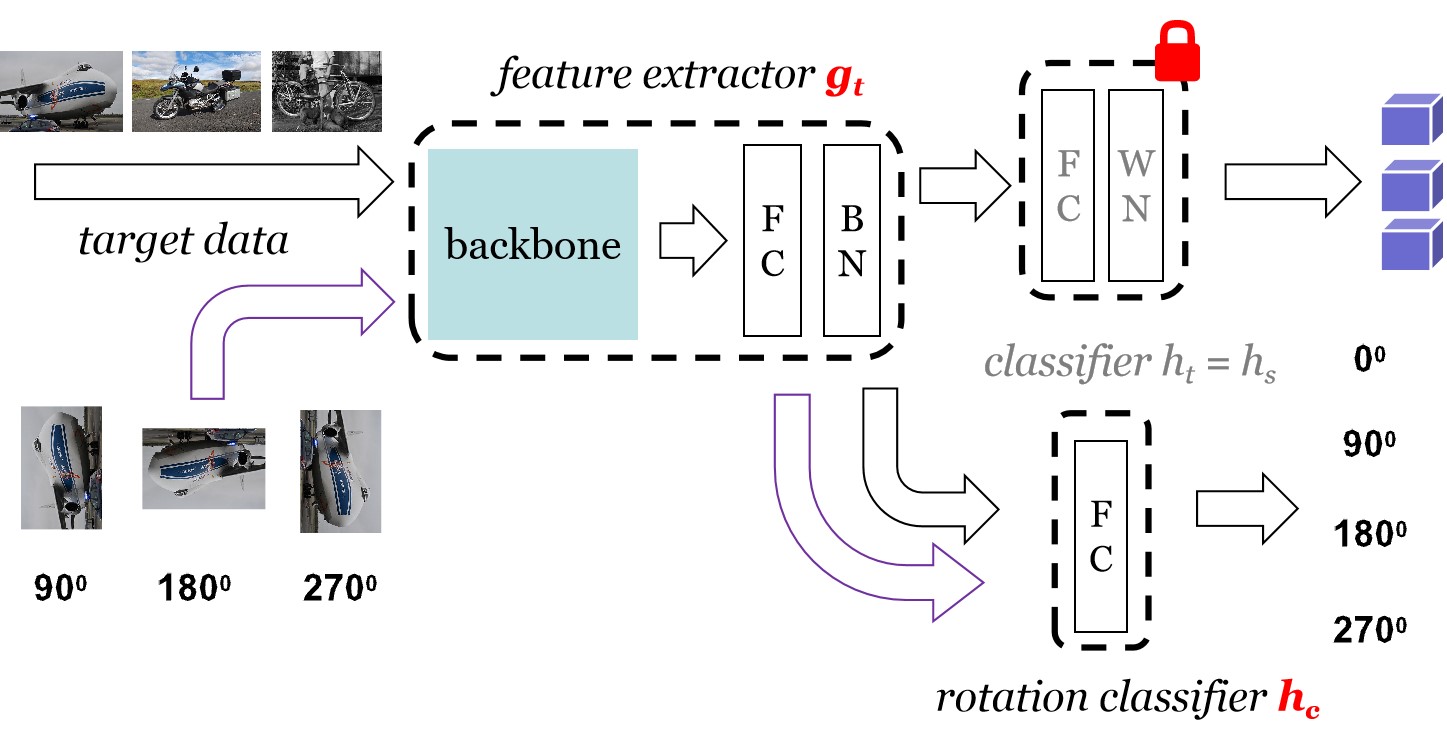}
		\caption{The pipeline of hypothesis transfer with self-supervised learning. Besides the common target model, we impose a rotation classifier $h_c$ after the feature encoding module $g_t$. $h_c$ is parameterized by a linear classifier, which aims to predict the \emph{relative} rotation of a target sample.} 
		\label{fig:framework_ssl}
	\end{figure}
	
	\begin{itemize}
		\item We first attain prototype representation (centriods) for each class in the target domain, similar to weighted k-means clustering,
		\begin{equation}
			c_k^{(0)} = \frac{\sum_{x\in \mathcal{X}_t}{\delta_k(\hat{f}_t(x))}\ \hat{g}_t(x)}{\sum_{x\in \mathcal{X}_t}{\delta_k(\hat{f}_t(x))}},
			\label{eq:proto}
		\end{equation}
		where $\delta_k(\cdot)$ denotes the $k$-th element in the
		soft-max output and $\hat{f}_t=\hat{g}_t \circ h_t$ denotes the previously learned target hypothesis. These centroids can robustly and more reliably characterize the distribution of different categories within the target domain.
		\item We then obtain new pseudo labels via the nearest centroid classifier:
		\begin{equation}
			\hat{y}_t = \arg\min_k D_f(\hat{g}_t(x), c_k^{(0)}),
			\label{eq:pseudo}
		\end{equation}
		where $D_f(a,b)$ measures the distance between $a$ and $b$.
		We use the cosine distance by default.
		\item Finally, we compute the target centroids based on the new pseudo labels:
		\begin{equation}
			\begin{aligned}
				c_k^{(1)} &= \frac{\sum_{x\in \mathcal{X}_t}{\mathds{1}(\hat{y}_t=k)}\ \hat{g}_t(x)}{\sum_{x\in \mathcal{X}_t}{\mathds{1}(\hat{y}_t=k)}},\\
				\hat{y}_t &= \arg\min_k D_f(\hat{g}_t(x), c_k^{(1)}).
			\end{aligned}
			\label{eq:proto2}
		\end{equation}
	\end{itemize} 
	
	We term $\hat{y}_t$ as self-supervised pseudo labels since they are generated by the centroids obtained in an unsupervised manner.
	Actually, this solution to pseudo labels behaves like that in Minimum Centroid Shift (MCS) \cite{liang2019distant} 
	where target-specific centroids and pseudo labels are alternately updated via optimizing the intra-class divergence minimization loss.
	In contrast, we employ the cross-entropy loss and just update the centroids and labels in Eq.~(\ref{eq:proto2}) for one round since updating once gives sufficiently good pseudo labels according to our observation in the experiment.
	We provide the cross-entropy loss of self-supervised pseudo-labeling below,
	\begin{equation}
		\begin{aligned}
			\mathcal{L}_{ssl}^{1} (f_t;\mathcal{X}_t,\hat{\mathcal{Y}}_t) = &\\
			-\gamma_1 \ \mathbb{E}_{(x,\hat{y}_t)\in \mathcal{X}_t \times \hat{\mathcal{Y}}_t} & \sum\nolimits_{k=1}^{K} \mathds{1}_{[k=\hat{y}_t]} \log \delta_k(f_t(x)),
		\end{aligned}
		\label{eq:ssl1}
	\end{equation} 
	where $\gamma_1>0$ is a regularization parameter for the trade-off between $\mathcal{L}_{ssl}^{1}$ and the main task in Eq.~(\ref{eq:im}).

	Also, we investigate the image-specific self-supervision in the unlabeled target domain. 
	Rotation prediction in RotNet \cite{gidaris2018unsupervised} aims to recognize one of four different 2d rotation (i.e., $0^{\circ}$, $90^{\circ}$, $180^{\circ}$, and $270^{\circ}$) that is applied to the image that it gets as input, which is a simple yet effective criterion in the self-supervised learning field.
	It is further verified by several recent studies \cite{wallace2020extending,mishra2021surprisingly} to learn semantically meaningful representations quite well, which is also desirable for domain adaptation problems.
	However, absolute rotation prediction is sensitive to some classification tasks.
	For example, in a main task aiming to distinguish digit `6' from digit `9', it is hard to determine which rotation category `9' belongs to, since `9' could also be a rotated `6' with 180 degrees or a rotated `9' with 0 degrees.
	To resolve this dilemma, we propose a new self-supervised learning task by predicting the relative rotation of each image pair.
	As shown in Fig.~\ref{fig:framework_ssl}, the relative rotation predictor is represented by $h_c:\mathbb{R}^{2d}\to \{1,2,3,4\}$ that takes the concatenated features of an image pair as input and maps them to one of four different rotation degrees.

	For an image in the target domain $x_i \in \mathcal{X}_t$, we first randomly sample an integral number $z_i$ from $[1,2,3,4]$ which corresponds to the rotation degree pool [$0^{\circ}$, $90^{\circ}$, $180^{\circ}$, $270^{\circ}$].
	Then we obtain the transformed image $x_i^{z_i}=Rot(x_i, z_i)$ by rotating $x_t$ with the associated degree $z_i$.
	Finally, the probability score of the $k$-th relative rotation degree predicted by $h_c$ is given by
	\begin{equation}
		\delta_k(h_c([g_t(x_i), g_t(x_i^{z_i})])),
		\label{eq:rot}
	\end{equation} 
	where $\delta_k(\cdot)$ denotes the $k$-th element in the soft-max output vector, and $[\cdot, \cdot]$ denotes the feature-level concatenation function.
	Therefore, the self-supervised rotation prediction loss is defined as
	\begin{equation}
		\begin{aligned}
			\mathcal{L}_{ssl}^{2} (g_t, h_c;\mathcal{X}_t,{\mathcal{Z}}_t) &= \\
			-\gamma_2 \ \mathbb{E}_{(x_i,\hat{z}_i)\in \mathcal{X}_t \times \mathcal{Z}_t} \sum\nolimits_{k=1}^{4} \mathds{1}_{[k=z_i]} &\log \delta_k(h_c([g_t(x_i), g_t(x_i^{z_i})])),
		\end{aligned}
		\label{eq:ssl2}
	\end{equation} 
	where $\gamma_2>0$ is a regularization parameter for the trade-off between $\mathcal{L}_{ssl}^{2}$ and the main loss, i.e. Eq.~(\ref{eq:im}).
	
	We provide an illustrative example of the complete hypothesis transfer framework in Fig.~\ref{fig:framework_ssl}.
	To summarize, given the source model $f_s=g_s\circ h_s$ and pseudo labels $\hat{\mathcal{Y}}_t$ generated in Eq.~(\ref{eq:proto2}) and randomly generated rotation labels $\mathcal{Z}_t$ as above, SHOT freezes the hypothesis from the source via $h_t\equiv h_s$ and learns the feature encoding module $g_t$ with the full optimization objective as 
	\begin{equation}
		\begin{aligned}
			&\mathcal{L}(g_t,h_c) = \mathcal{L}_{ent} (h_t\circ g_t;\mathcal{X}_t) + \beta \mathcal{L}_{div}(h_t\circ g_t;\mathcal{X}_t) \ - \\
			& \gamma_1 \ \mathbb{E}_{(x,\hat{y}_t)\in \mathcal{X}_t \times \hat{\mathcal{Y}}_t} \sum\nolimits_{k=1}^{K} \mathds{1}_{[k=\hat{y}_t]} \log \delta_k(f_t(x)) \ - \\
			& \gamma_2 \ \mathbb{E}_{(x_i,\hat{z}_i)\in \mathcal{X}_t \times \mathcal{Z}_t} \sum\nolimits_{k=1}^{4} \mathds{1}_{[k=z_i]} \log \delta_k(h_c([g_t(x_i), g_t(x_i^{z_i})])).
		\end{aligned}
		\label{eq:shot}
	\end{equation} 
	
	\subsection{Labeling Transfer with Semi-supervised Learning}
	\label{sec:mixmatch}
	After we obtain the predictions for all the samples in the target domain via SHOT in Eq.~(\ref{eq:shot}), we can measure the confidence scores of these predictions via the entropy function $\mathbb{H}(p)=\sum_i p_i\log p_i$, where $p$ is a probability prediction vector.
	Observing the distribution of confidence scores, we find that there always exist some less confident (high-entropy) predictions that are possibly inaccurate.
	Fortunately, we can utilize the reliable labeling information from high confident predictions to improve the accuracy of these less confident ones.
	To this end, we propose a two-step method to enforce the information propagation from low-entropy predictions to high-entropy ones.
	In the first step, we divide the target domain into two splits according to the confidence scores and treat these two splits as a labeled subset and an unlabeled subset, respectively.
	In the second step, we readily employ a semi-supervised learning algorithm to learn the enhanced predictions for the unlabeled set here.
	
	\begin{figure}[!t]
		\centering
		\includegraphics[width=0.45\textwidth]{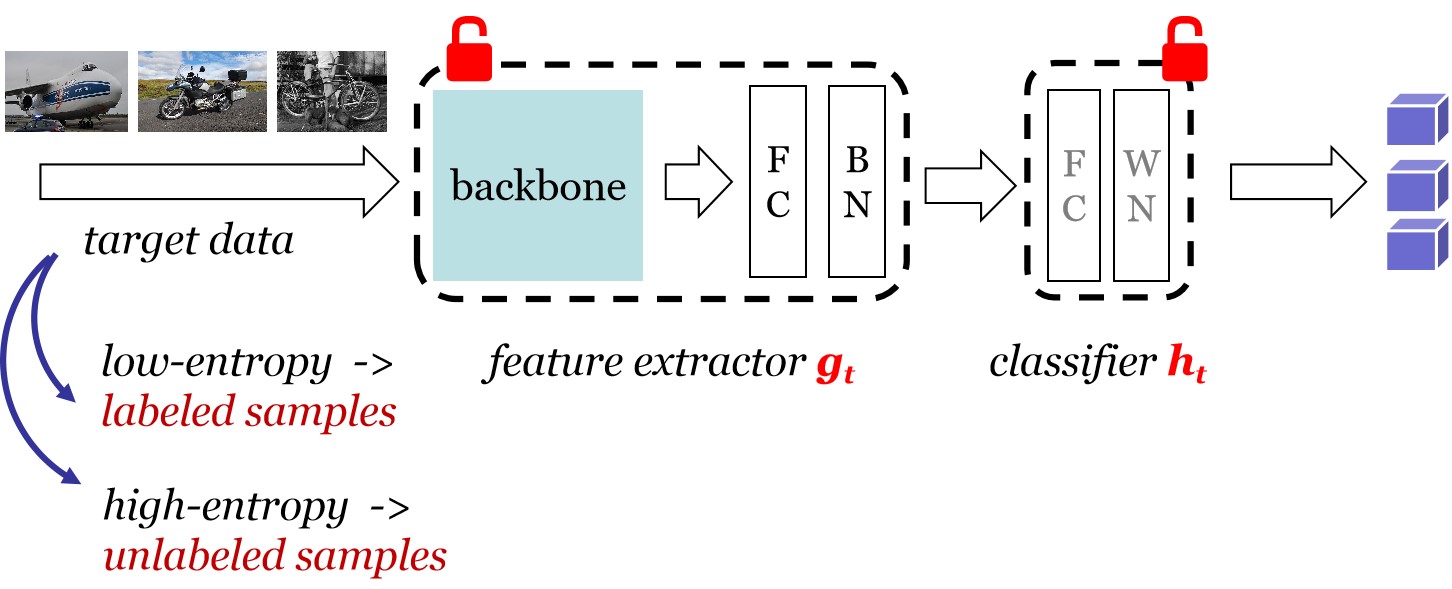}
		\caption{The pipeline of the labeling transfer strategy with semi-supervised learning. Both the feature encoding module $g_t$ and the classification module $h_t$ are learned via the MixMatch~\cite{berthelot2019mixmatch} algorithm.} 
		\label{fig:framework_plus}
	\end{figure}
	
	Regarding the choice of a semi-supervised learning algorithm in the second step, we simply adopt a popular and well-performing approach, MixMatch \cite{berthelot2019mixmatch}, which unifies consistency regularization, entropy minimization, and the MixUp regularization into one framework.
	Then the key point lies in \emph{how to divide the target domain into two splits.}
	With average entropy, we first obtain the proportion of the labeled subset in the entire target domain by automatically computing
	\begin{equation}
		a = \frac{\sum_i \mathds{1}(\xi_i < \frac{\sum_j \xi_j}{n_t})}{n_t},
	\end{equation}
	where $\xi \in \mathcal{R}^{n_t}$ denotes the entropy values of all the predictions in the target domain, where $\xi_i = \mathbb{H}(\delta(f_t(x_i))), i \in [1, \cdots, n_t]$.
	Then for each class $k \in [0, K]$, we put the index with entropy values among the top $t_k$ smallest into the index pool of labeled split, where
	\begin{equation}
		t_k = \lfloor a \sum\nolimits_i \mathds{1}(\bar{y}_i=k) \rfloor, \ k \in [1, \cdots, K],
	\end{equation}
	and $\bar{y}_i \in [1,K]^{n_t}$ is the predicted label by SHOT in Eq.~(\ref{eq:shot}).
	In this manner, we get the labeled split, and the remaining samples constitute the unlabeled split.
	We call this strategy in Fig.~\ref{fig:framework_plus} as labeling transfer since in this stage we only need the labeling information (predictions) while the feature encoding module $g_t$ is initialized with that learned in Eq.~(\ref{eq:shot}).
	Besides, the classification module $h_t$ is newly initialized from scratch and not frozen any more.
	So far, we develop a two-stage approach, called SHOT++, in which the first stage is SHOT in Eq.~(\ref{eq:shot}) and the second stage is the proposed labeling transfer strategy in Fig.~\ref{fig:framework_plus}.
	
	\subsection{Extension to Multi-source Domain Adaptation}
	We also provide an extension of the proposed SHOT approach for multi-source domain adaptation (MSDA) \cite{peng2019moment}.
	Different from vanilla UDA (one source and one target), there are multiple sources in the MSDA task.
	For simplicity, we run SHOT and SHOT-IM on each source-target pair and then sum up the probabilistic scores obtained from each pair.
	Finally, we get the predictions of samples in the target domain via the \emph{argmax} operation.
	As for labeling transfer, we split the target domain into two pieces for each pair, and learn the independent prediction scores.
	
	\subsection{Extension to Partial-set Domain Adaptation}
	\label{sec:pda}
	We also provide an extension of the proposed SHOT approach for partial-set domain adaptation (PDA) \cite{cao2018partiala}.
	PDA differs from vanilla UDA in that the target label space is a subset of that of the source label space.
	Looking at the diversity-promoting term $\mathcal{L}_{div}$ in Eq.~(\ref{eq:shot}), it encourages the target domain to own a uniform label distribution.
	Though seemingly reasonable for solving closed-set UDA, it is not suitable for PDA.
	In reality, the target domain only contains some classes of all the classes in the source domain, making the label distribution sparse.
	Hence, we drop the second term $\mathcal{L}_{div}$ for PDA by letting $\beta=0$.
	
	Besides, within the self-supervised pseudo-labeling strategy, we usually need to obtain $K$ centroids in the target domain. 
	However, for the PDA task, there are some tiny centroids which should be considered as empty like in k-means clustering.
	Therefore, SHOT discards tiny centroids with size smaller than $T_{c}$ in Eq.~(\ref{eq:proto2}) for PDA problems. 
	
	\subsection{Extension to Semi-supervised Domain Adaptation}
	We further extend the proposed SHOT approach for semi-supervised domain adaptation (SSDA) \cite{saito2019semi}.
	SSDA differs from UDA in that some labeled data exist in the target domain.
	Therefore, we adopt the supervised training loss in Eq.~(\ref{eq:ls}) for labeled target data and the complete loss in Eq.~(\ref{eq:shot}) for unlabeled target data.
	Besides, we also consider the labeled target data when computing the target-specific centriods.
	As for labeling transfer, we split the unlabeled target domain into two pieces and then add the labeled data into the labeled split.

	\subsection{Network Architecture}
	\label{sec:network}
	Here we discuss some architecture choices for the neural network model to parameterize both the feature encoding module and the hypothesis. 
	First, we need to look back at the expected network outputs for cross-entropy loss in Eq.~(\ref{eq:cross}).
	If $y_s=k$, then maximizing $f_s^{(k)}(x_s)=\frac{\exp(w_k^\top g_s(x_s))}{\sum_i \exp(w_i^\top g_s(x_s))}$ means minimizing the distance between $g_s(x_s)$ and $w_k$, where $w_k$ is the $k$-th weight vector in the last FC layer.
	Ideally, all the samples from the $k$-th class would have a feature embedding near to $w_k$. 
	If unlabeled target samples are given the correct pseudo labels, it is easily understandable that source feature embeddings are similar to target ones via the pseudo-labeling term in Eq.~(\ref{eq:ssl1}).
	The intuition behind is quite similar to previous studies \cite{long2013transfer,xie2018learning} where a simplified MMD is exploited for multi-modal domain confusion.
	Since the weight norm matters in the inner distance within the soft-max output, we adopt weight normalization (WN) \cite{salimans2016weight} to keep the norm of each weight vector $w_i$ the same in the FC classifier layer.
	Besides, as indicated in prior studies, batch normalization (BN) \cite{ioffe2015batch} can reduce the internal dataset shift since different domains share the same mean (zero) and variance which can be considered as first-order and second-order moments. 
	Based on these considerations, we form the frameworks of SHOT and SHOT++ as shown in Figs.~\ref{fig:framework}$\sim$\ref{fig:framework_plus}. 
	
	\section{Experiments}
	\subsection{Setup}
	To testify their versatility, we evaluate our methods in three unsupervised DA scenarios (i.e. closed-set, partial-set, multi-source), and one semi-supervised DA scenario over several popular visual benchmarks as introduced below.
	
	\textbf{Digits} is a widely used DA benchmark that focuses on digit recognition. We follow the protocol of \cite{long2018conditional} and utilize three representative subsets: SVHN (\textbf{S}), MNIST (\textbf{M}), and USPS (\textbf{U}).
	We train our model using the training sets of each domain and report the recognition results on the standard test set of the target domain.

	\textbf{Office} \cite{saenko2010adapting} is a standard DA benchmark which contains three domains, i.e., Amazon (\textbf{A}), DSLR (\textbf{D}), and Webcam (\textbf{W}), and each domain includes 31 object classes in the office environment. 
	Gong et al.~\cite{gong2012geodesic} further extract 10 shared categories between Office and Caltech-256 (\textbf{C}) to form a new benchmark named \textbf{Office-Caltech}. 
	Both \textbf{Office} and \textbf{Office-Caltech} are considered small-sized.
	
	\textbf{Office-Home} \cite{venkateswara2017deep} is a challenging medium-sized benchmark, which consists of four distinct domains, i.e., Artistic images (\textbf{Ar}), Clip Art (\textbf{Cl}), Product images (\textbf{Pr}), and Real-World images (\textbf{Rw}). There are totally 65 everyday object categories in each domain.
	
	\textbf{VisDA-C} \cite{peng2017visda} is a challenging large-scale benchmark that mainly focuses on the 12-class synthesis-to-real object recognition task. The source domain contains 152 thousand synthetic (\textbf{S}) images generated by rendering 3D models while the target domain has 55 thousand real (\textbf{R}) object images sampled from Microsoft COCO.
	
	\textbf{PACS}~\cite{li2017deeper} is a popular benchmark for multi-source domain adaptation. It contains four different domains, i.e., Art painting (\textbf{A}), Cartoon (\textbf{C}), Photo (\textbf{P}), and Sketch (\textbf{S}). There are totally 7 common categories in each domain.
	
	\textbf{Baseline methods.}
	For vanilla unsupervised DA in digit recognition, we compare SHOT with ADDA~\cite{tzeng2017adversarial}, ADR~\cite{saito2018adversarial}, CDAN~\cite{long2018conditional}, CyCADA~\cite{hoffman2018cycada}, CAT~\cite{deng2019cluster}, SWD~\cite{lee2019sliced} and STAR~\cite{lu2020stochastic};
	for object recognition, we compare ours with DANN~\cite{ganin2015unsupervised}, DAN~\cite{long2015learning}, SAFN~\cite{xu2019larger}, BSP~\cite{chen2019transferability}, MDD~\cite{zhang2019bridging}, TransNorm~\cite{wang2019transferable}, DSBN~\cite{chang2019domain}, BNM~\cite{cui2020towards} and GVB-GD~\cite{cui2020gradually}.
	For partial-set DA tasks, we compare ours with IWAN~\cite{zhang2018importance}, SAN~\cite{cao2018partiala}, ETN~\cite{cao2019learning}, DRCN~\cite{li2020deep_pda}, RTNet$_{adv}$~\cite{chen2020selective}, BA$^3$US~\cite{liang2020balanced}, and TSCDA~\cite{ren2020learning}.
	For multi-source UDA, we compare ours with DCTN~\cite{xu2018deep}, MCD~\cite{saito2018maximum},
	WBN~\cite{mancini2018boosting},
	M$^3$SDA-$\beta$ \cite{peng2019moment}, 
	Meta-MCD~\cite{li2020online},
	SImAl~\cite{venkat2020your},
	and CMSS~\cite{yang2020curriculum}.
	For SSDA, we mainly compare our methods with MME~\cite{saito2019semi} and UODA~\cite{qin2021opposite}.
	Note that results are directly cited from published papers if we follow the same setting.
	`Source-model-only' (also called `src-only') denotes using the entire model learned from the source domain for target label prediction.
	`labeled-data-only' denotes using labeled target data only when learning the feature extractor $g_t$.
	`Target-supervised' denotes training with the target data itself. For datasets without train-validation splits, we divide the target domain into three parts (\textbf{0.6}/\textbf{0.2}/\textbf{0.2}) as training, validation, and testing sets. Then we train the network via the training set and the validation set and finally report the accuracy on the testing set.
	SHOT-IM is a special case of SHOT, where both self-supervised losses are ignored by letting $\gamma_1=\gamma_2=0$ in Eq.~(\ref{eq:shot}).

	\subsection{Implementation Details}
	\textbf{Network architecture.}
	For the digit recognition task, we use the same architectures with CDAN~\cite{long2018conditional}, namely, the classical LeNet-5 \cite{lecun1998gradient} network for USPS$\leftrightarrow$MNIST and a variant of LeNet for SVHN$\to$MNIST. 
	More network details can be found in \emph{Appendix A} of \cite{liang2020we}.
	For the object recognition task, we employ the pre-trained ResNet-50 or ResNet-101 \cite{he2016deep} models as the backbone, like \cite{long2018conditional,deng2019cluster,xu2019larger,peng2019moment}.
	Following \cite{ganin2015unsupervised}, we replace the original FC layer with a bottleneck layer (256 units) and a task-specific FC classifier layer in Fig.~\ref{fig:framework}.
	Precisely, a BN layer is put after FC inside the bottleneck layer and a weight normalization layer is utilized in the task-specific FC layer.
	
	\textbf{Network hyper-parameters.}
	We train the whole network through back-propagation, and the newly added layers are trained with a learning rate 10 times that of the pre-trained layers (backbone shown in Fig.~\ref{fig:framework}).
	Concretely, we adopt mini-batch SGD with momentum 0.9, weight decay 1e$^{-3}$ and learning rate $\eta_0=1e^{-2}$ for the new layers and those layers learned from scratch for all experiments except $\eta_0=1e^{-3}$ for \textbf{VisDA-C}.
	We further adopt the same learning rate scheduler $\eta=\eta_0 \cdot (1+10\cdot p)^{-0.75}$ as \cite{ganin2015unsupervised,long2018conditional}, where $p$ is the training progress changing from 0 to 1.
	Besides, we set the batch size to 64 for all the tasks. 
	We utilize $\gamma_1=0.3,\gamma_2=0.6$ for all experiments except $\gamma_1=0.1,\gamma_2=0.2$ for \textbf{Digits} in Table~\ref{table:digit} and SSDA in Table~\ref{tab:ssda}.
	Concerning the labeling transfer strategy, only `source-model-only++' for object recognition does not use the learned source model as initialization.

	For \textbf{Digits}, we train the best source hypothesis using the test set of the source dataset as validation.
	For other datasets without train-validation splits, we randomly specify a \textbf{0.9/ 0.1} split in the source dataset and generate the best source hypothesis based on the validation split.
	The maximum number of epochs for \textbf{Digits}, \textbf{Office}, \textbf{Office-Home}, \textbf{VisDA-C} and \textbf{Office-Caltech} is empirically set as 30, 100, 50, 10, and 100, respectively.
	For learning in the target domain, we update the pseudo-labels epoch by epoch, and the maximum number of epochs is empirically set as 15.
	Regarding the second step in Section \ref{sec:mixmatch}, we adopt the same learning setting as that of training SHOT and the default parameters $\alpha=0.75$ within MixMatch~\cite{berthelot2019mixmatch}.
	We utilize $\alpha=0.1$ only for \textbf{Digits}.
	Besides, we randomly run our methods for three times with different random seeds $\{2019, 2020, 2021\}$ via \textbf{PyTorch}, and report the mean accuracy.
	Note that we do not use any target augmentation such as the ten-crop ensemble \cite{long2018conditional} for evaluation.
	
	\setlength{\tabcolsep}{3.0pt}
	\begin{table}[tbp]
		\centering
		\small
		\caption{Classification accuracies (\%) on \textbf{Digits} dataset for \emph{vanilla closed-set UDA}. S: SVHN, M:MNIST, U: USPS. (Best value is in \textbf{\color{blush}red color})}
		\label{table:digit}
		\vskip -0.05in
		\resizebox{0.48\textwidth}{!}{$
			\begin{tabular}{lccca}
				\toprule
				Method (Source $\to$ Target) & S $\to$ M & U $\to$ M & M $\to$ U & Avg.\\
				\midrule
				Source only \cite{hoffman2018cycada} & 67.1$\pm$0.6 & 69.6$\pm$3.8 & 82.2$\pm$0.8 & 73.0 \\
				ADDA \cite{tzeng2017adversarial} & 76.0$\pm$1.8 & 90.1$\pm$0.8 & 89.4$\pm$0.2 & 85.2 \\
				ADR \cite{saito2018adversarial} & 95.0$\pm$1.9 & 93.1$\pm$1.3 & 93.2$\pm$2.5 & 93.8 \\
				CyCADA \cite{hoffman2018cycada} & 90.4$\pm$0.4 & 96.5$\pm$0.1 & 95.6$\pm$0.4 & 94.2 \\
				CDAN \cite{long2018conditional} & \multicolumn{1}{l}{89.2} & \multicolumn{1}{l}{\textbf{\color{blush}98.0}} & \multicolumn{1}{l}{95.6} & 94.3 \\
				rRevGrad+CAT \cite{deng2019cluster} & 98.8$\pm$0.0 & 96.0$\pm$0.9 & 94.0$\pm$0.7 & 96.3 \\
				SWD \cite{lee2019sliced} & 98.9$\pm$0.1 & 97.1$\pm$0.1 & 98.1$\pm$0.1 & 98.0 \\
				STAR \cite{lu2020stochastic} & 98.8$\pm$0.1 & 97.7$\pm$0.1 & 97.8$\pm$0.1 & 98.1 \\
				\midrule
				Source-model-only & 71.2$\pm$0.7 & 88.0$\pm$2.5 & 78.4$\pm$2.3 & 79.2 \\
				SHOT-IM & 98.5$\pm$0.8 & 97.6$\pm$0.1 & 97.8$\pm$0.4 & 97.9 \\
				SHOT & \textbf{\color{blush}99.0}$\pm$0.1 & 97.6$\pm$0.2 & 97.8$\pm$0.3 & 98.1 \\
				\midrule
				Source-model-only++ & 88.7$\pm$3.0 & 94.5$\pm$0.9 & 89.8$\pm$2.0 & 91.0 \\
				SHOT-IM++ & 98.5$\pm$0.8 & 97.7$\pm$0.1 & \textbf{\color{blush}98.4}$\pm$0.4 & 98.2 \\
				SHOT++ & 98.9$\pm$0.1 & 97.8$\pm$0.1 & \textbf{\color{blush}98.4}$\pm$0.1 & \textbf{\color{blush}98.4} \\
				\midrule
				Target-supervised & 99.2$\pm$0.1 & 99.2$\pm$0.1 & 96.8$\pm$0.2 & 98.4 \\
				\bottomrule
			\end{tabular}
			$}
	\end{table}

	\setlength{\tabcolsep}{1.0pt}
	\begin{table}[tbp]
	\centering
	\small
	\caption{Classification accuracies (\%) on small-sized \textbf{Office} dataset for \emph{vanilla closed-set UDA} (ResNet-50).}
	\label{table:office}
	\vskip -0.05in
	\resizebox{0.48\textwidth}{!}{$
		\begin{tabular}{lccccccacccccca}
			\toprule
			Method (Source$\to$Target) & A$\to$D & A$\to$W & D$\to$A & D$\to$W & W$\to$A & W$\to$D & Avg. \\
			\midrule
			ResNet-50 \cite{he2016deep}  & 68.9 & 68.4 & 62.5 & 96.7 & 60.7 & 99.3 & 76.1 \\
			DAN \cite{long2015learning}  & 78.6 & 80.5 & 63.6 & 97.1 & 62.8 & 99.6 & 80.4 \\
			DANN \cite{ganin2015unsupervised} & 79.7 & 82.0 & 68.2 & 96.9 & 67.4 & 99.1 & 82.2 \\
			SAFN+ENT \cite{xu2019larger} & 90.7 & 90.1 & 73.0 & 98.6 & 70.2 & 99.8 & 87.1 \\
			rRevGrad+CAT \cite{deng2019cluster} & 90.8 & 94.4 & 72.2 & 98.0 & 70.2 & \textbf{\color{blush}100.} & 87.6 \\
			CDAN \cite{long2018conditional} & 92.9 & 94.1 & 71.0 & 98.6 & 69.3 & \textbf{\color{blush}100.} & 87.7 \\
			DSBN+MSTN \cite{chang2019domain} & 92.2 & 92.7 & 71.7 & \textbf{\color{blush}99.0} & 74.4 & \textbf{\color{blush}100.} & 88.3 \\
			CDAN+BSP \cite{chen2019transferability} & 93.0 & 93.3 & 73.6 & 98.2 & 72.6 & \textbf{\color{blush}100.} & 88.5 \\
			CDAN+BNM \cite{cui2020towards} & 92.9 & 92.8 & 73.5 & 98.8 & 73.8 & \textbf{\color{blush}100.} & 88.6 \\
			MDD~\cite{zhang2019bridging} & 93.5 & 94.5 & 74.6 & 98.4 & 72.2 & \textbf{\color{blush}100.} & 88.9 \\
			CDAN+TransNorm \cite{wang2019transferable} & 94.0 & \textbf{\color{blush}95.7} & 73.4 & 98.7 & 74.2 & \textbf{\color{blush}100.} & \textbf{\color{blush}89.3} \\
			GVB-GD~\cite{cui2020gradually} & \textbf{\color{blush}95.0} & 94.8 & 73.4 & 98.7 & 73.7 & \textbf{\color{blush}100.} & \textbf{\color{blush}89.3} \\
			\midrule
			Source-model-only & 
			80.2 & 76.9 & 60.3 & 95.4 & 63.6 & 98.9 & 79.2 \\
			SHOT-IM & 90.2 & 91.1 & 72.4 & 98.3 & 71.8 & 99.9 & 87.3 \\
			SHOT & 93.9 & 90.1 & 75.3 & 98.7 & 75.0 & 99.9 & 88.8 \\
			\midrule
			Source-model-only++ & 88.5 & 87.3 & 69.0 & 97.7 & 70.8 & 99.0 & 85.4 \\			
			SHOT-IM++ & 90.9 & 91.9 & 73.5 & 98.6 & 72.5 & 99.7 & 87.8 \\
			SHOT++ & 94.3 & 90.4 & \textbf{\color{blush}76.2} & 98.7 & \textbf{\color{blush}75.8} & 99.9 & 89.2 \\
			\midrule
			Target-supervised & 98.0 & 98.7 & 86.0 & 98.7 & 86.0 & 98.0 & 94.3 \\
			\bottomrule
		\end{tabular}
		$}
	\end{table}

	\setlength{\tabcolsep}{2.0pt}
	\begin{table*}[htbp]
	\centering
	\small
	\caption{Classification accuracies (\%) on medium-sized \textbf{Office-Home} dataset for \emph{vanilla closed-set UDA} (ResNet-50).}
	\label{table:home}
	\vskip -0.05in
	\resizebox{0.9\textwidth}{!}{$
		\begin{tabular}{lcccccccccccca}
			\toprule
			Method (Source$\to$Target) & Ar$\to$Cl & Ar$\to$Pr & Ar$\to$Re & Cl$\to$Ar & Cl$\to$Pr & Cl$\to$Re & Pr$\to$Ar & Pr$\to$Cl & Pr$\to$Re & Re$\to$Ar & Re$\to$Cl & Re$\to$Pr & Avg. \\
			\midrule
			ResNet-50 \cite{he2016deep}  & 34.9 & 50.0 & 58.0 & 37.4 & 41.9 & 46.2 & 38.5 & 31.2 & 60.4 & 53.9 & 41.2 & 59.9 & 46.1 \\
			DANN \cite{ganin2015unsupervised} & 45.6 & 59.3 & 70.1 & 47.0 & 58.5 & 60.9 & 46.1 & 43.7 & 68.5 & 63.2 & 51.8 & 76.8 & 57.6 \\
			DAN \cite{long2015learning} & 43.6 & 57.0 & 67.9 & 45.8 & 56.5 & 60.4 & 44.0 & 43.6 & 67.7 & 63.1 & 51.5 & 74.3 & 56.3 \\
			CDAN \cite{long2018conditional}  & 50.7 & 70.6 & 76.0 & 57.6 & 70.0 & 70.0 & 57.4 & 50.9 & 77.3 & 70.9 & 56.7 & 81.6 & 65.8 \\
			CDAN+BSP \cite{chen2019transferability} & 52.0 & 68.6 & 76.1 & 58.0 & 70.3 & 70.2 & 58.6 & 50.2 & 77.6 & 72.2 & 59.3 & 81.9 & 66.3 \\
			SAFN \cite{xu2019larger}  & 52.0 & 71.7 & 76.3 & 64.2 & 69.9 & 71.9 & 63.7 & 51.4 & 77.1 & 70.9 & 57.1 & 81.5 & 67.3 \\
			CDAN+TransNorm \cite{wang2019transferable} & 50.2 & 71.4 & 77.4 & 59.3 & 72.7 & 73.1 & 61.0 & 53.1 & 79.5 & 71.9 & 59.0 & 82.9 & 67.6 \\
			MDD~\cite{zhang2019bridging} & 54.9 & 73.7 & 77.8 & 60.0 & 71.4 & 71.8 & 61.2 & 53.6 & 78.1 & 72.5 & 60.2 & 82.3 & 68.1 \\
			CDAN+BNM \cite{cui2020towards} & 56.2 & 73.7 & 79.0 & 63.1 & 73.6 & 74.0 & 62.4 & 54.8 & 80.7 & 72.4 & 58.9 & 83.5 & 69.4 \\
			GVB-GD~\cite{cui2020gradually} & 57.0 & 74.7 & 79.8 & 64.6 & 74.1 & 74.6 & 65.2 & 55.1 & 81.0 & \textbf{\color{blush}74.6} & 59.7 & 84.3 & 70.4 \\
			\midrule
			Source-model-only & 44.5 & 67.6 & 74.7 & 52.5 & 62.8 & 64.9 & 53.1 & 40.5 & 73.3 & 65.3 & 45.1 & 77.9 & 60.2 \\
			SHOT-IM & 55.9 & 76.8 & 80.6 & 66.7 & 73.7 & 75.4 & 65.4 & 54.9 & 80.9 & 73.2 & 58.5 & 83.5 & 70.5 \\
			SHOT & 57.7 & 79.1 & 81.5 & 67.6 & 77.9 & 77.8 & 68.1 & 55.8 & 82.0 & 72.8 & 59.7 & 84.4 & 72.0 \\
			\midrule
			Source-model-only++ & 50.2 & 75.9 & 79.7 & 62.6 & 74.3 & 74.8 & 59.5 & 44.4 & 79.9 & 69.4 & 45.4 & 83.1 & 66.6 \\
			SHOT-IM++ & 56.9 & 77.7 & 81.5 & 67.6 & 74.9 & 76.9 & 66.1 & 55.9 & 81.7 & 73.8 & 59.3 & 84.4 & 71.4 \\
			SHOT++ & \textbf{\color{blush}57.9} & \textbf{\color{blush}79.7} & \textbf{\color{blush}82.5} & \textbf{\color{blush}68.5} & \textbf{\color{blush}79.6} & \textbf{\color{blush}79.3} & \textbf{\color{blush}68.5} & \textbf{\color{blush}57.0} & \textbf{\color{blush}83.0} & 73.7 & \textbf{\color{blush}60.7} & \textbf{\color{blush}84.9} & \textbf{\color{blush}73.0} \\
			\midrule
			Target-supervised & 77.9 & 91.4 & 84.4 & 74.5 & 91.4 & 84.4 & 74.5 & 77.9 & 84.4 & 74.5 & 77.9 & 91.4 & 82.0 \\
			\bottomrule
		\end{tabular}
		$}
	\end{table*}

	\setlength{\tabcolsep}{4.0pt}
	\begin{table*}[htbp]
	\centering
	\small
	\caption{Classification accuracies (\%) on large-scale \textbf{VisDA-C} dataset for \emph{vanilla closed-set UDA} (ResNet-101).}
	\label{table:visda}
	\vskip -0.05in
	\resizebox{0.9\textwidth}{!}{$
		\begin{tabular}{lcccccccccccca}
			\toprule
			Method (Synthesis $\to$ Real) & plane & bcycl & bus & car & horse & knife & mcycl & person & plant & sktbrd & train & truck & Per-class \\
			\midrule
			ResNet-101 \cite{he2016deep} & 55.1 & 53.3 & 61.9 & 59.1 & 80.6 & 17.9 & 79.7 & 31.2 & 81.0 & 26.5 & 73.5 & 8.5 & 52.4 \\
			DANN \cite{ganin2015unsupervised} & 81.9 & 77.7 & 82.8 & 44.3 & 81.2 & 29.5 & 65.1 & 28.6 & 51.9 & 54.6 & 82.8 & 7.8 & 57.4 \\
			DAN \cite{long2015learning} & 87.1 & 63.0 & 76.5 & 42.0 & 90.3 & 42.9 & 85.9 & 53.1 & 49.7 & 36.3 & 85.8 & 20.7 & 61.1 \\
			ADR \cite{saito2018adversarial} & 94.2 & 48.5 & 84.0 & 72.9 & 90.1 & 74.2 & 92.6 & 72.5 & 80.8 & 61.8 & 82.2 & 28.8 & 73.5 \\
			CDAN+BSP \cite{chen2019transferability} & 92.4 & 61.0 & 81.0 & 57.5 & 89.0 & 80.6 & 90.1 & 77.0 & 84.2 & 77.9 & 82.1 & 38.4 & 75.9 \\
			SAFN \cite{xu2019larger} & 93.6 & 61.3 & 84.1 & 70.6 & 94.1 & 79.0 & 91.8 & 79.6 & 89.9 & 55.6 & 89.0 & 24.4 & 76.1 \\
			SWD \cite{lee2019sliced} & 90.8 & 82.5 & 81.7 & 70.5 & 91.7 & 69.5 & 86.3 & 77.5 & 87.4 & 63.6 & 85.6 & 29.2 & 76.4 \\
			DSBN+MSTN \cite{chang2019domain} & 94.7 & 86.7 & 76.0 & 72.0 & 95.2 & 75.1 & 87.9 & 81.3 & 91.1 & 68.9 & 88.3 & 45.5 & 80.2 \\
			DTA \cite{lee2019drop} & 93.7 & 82.2 & 85.6 & 83.8 & 93.0 & 81.0 & 90.7 & 82.1 & 95.1 & 78.1 & 86.4 & 32.1 & 81.5 \\
			STAR \cite{lu2020stochastic} & 95.0 & 84.0 & 84.6 & 73.0 & 91.6 & 91.8 & 85.9 & 78.4 & 94.4 & 84.7 & 87.0 & 42.2 & 82.7 \\
			CAN \cite{kang2019contrastive} & 97.0 & 87.2 & 82.5 & 74.3 & 97.8 & 96.2 & 90.8 & 80.7 & 96.6 & \textbf{\color{blush}96.3} & 87.5 & \textbf{\color{blush}59.9} & 87.2 \\
			\midrule
			Source-model-only & 64.1 & 24.9 & 53.0 & 66.5 & 67.9 & 9.1 & 84.5 & 21.1 & 62.8 & 29.8 & 83.5 & 9.3 & 48.0 \\
			SHOT-IM & 93.7 & 86.4 & 78.7 & 50.6 & 91.0 & 93.6 & 79.0 & 78.3 & 89.3 & 85.4 & 88.0 & 51.1 & 80.4 \\
			SHOT & 95.8 & 88.2 & 87.2 & 73.7 & 95.2 & 96.4 & 87.9 & \textbf{\color{blush}84.5} & 92.5 & 89.3 & 85.7 & 49.1 & 85.5 \\
			\midrule
			Source-model-only++ & 73.0 & 12.9 & 76.1 & \textbf{\color{blush}90.3} & 93.7 & 1.5 & \textbf{\color{blush}94.9} & 40.9 & 84.6 & 75.1 & 91.2 & 4.9 & 61.6 \\
			SHOT-IM++ & 96.7 & 87.6 & 89.2 & 71.4 & 96.3 & 98.5 & 91.9 & 79.9 & 95.5 & 86.4 & 93.5 & 32.9 & 85.0 \\
			SHOT++ & \textbf{\color{blush}97.7} & \textbf{\color{blush}88.4} & \textbf{\color{blush}90.2} & 86.3 & \textbf{\color{blush}97.9} & \textbf{\color{blush}98.6} & \textbf{\color{blush}92.9} & 84.1 & \textbf{\color{blush}97.1} & 92.2 & \textbf{\color{blush}93.6} & 28.8 & \textbf{\color{blush}87.3} \\
			\midrule
			Target-supervised & 97.0 & 86.6 & 84.3 & 88.7 & 96.3 & 94.4 & 92.0 & 89.4 & 95.5 & 91.8 & 90.7 & 68.7 & 89.6 \\
			\bottomrule
		\end{tabular}
		$}
	\end{table*}

	\subsection{Results of Digit Recognition (Vanilla Closed-set)}
	For digit recognition, we evaluate our methods on three popular closed-set unsupervised domain adaptation tasks, i.e., SVHN$\to$MNIST, USPS$\to$MNIST, and MNIST$\to$USPS.
	The classification accuracies of our methods and prior work are reported in Table~\ref{table:digit}.
	Obviously, SHOT obtains the best mean accuracy for each task and also outperforms prior work in terms of the average accuracy.
	Compared with the baseline method source-model-only, SHOT-IM always achieves better results, and SHOT performs better than SHOT-IM due to the contribution of self-supervised learning in the target domain.
	Taking into consideration the labeling transfer strategy, all three methods are able to obtain enhanced classification results, indicating the effectiveness of intra-domain semi-supervised learning.
	It is also worth noting that SHOT++ even
	offers superior performance to the target-supervised result in MNIST$\to$USPS. 
	This may be because MNIST is much larger than USPS, which alleviates the domain shift well. 
	
	\subsection{Results of Object Recognition (Vanilla Closed-set)}
	Next, we evaluate our methods on object recognition benchmarks including \textbf{Office}, \textbf{Office-Home} and \textbf{VisDA-C} under the vanilla closed-set DA setting.
	As shown in Table~\ref{table:office}, SHOT performs the best for two challenging tasks, D$\to$A and W$\to$A, and obtains an average accuracy 88.8\% that is competitive to two state-of-the-art methods, MDD~\cite{zhang2019bridging} and BNM~\cite{cui2020towards}.
	Similar to the observations in Table~\ref{table:digit}, the labeling transfer strategy is beneficial to cross-domain object recognition, and SHOT++ obtains the same mean accuracy as previous state-of-the-art methods, TransNorm~\cite{wang2019transferable} and GVB-GD~\cite{cui2020gradually}.
	This may be because SHOT needs a relatively large target domain to learn the target-specific module $g_t$ while $D$ and $W$ as the target domain are not big enough.
	Generally, SHOT obtains competitive performance even with no direct access to the source domain data.
	
	\setlength{\tabcolsep}{2.5pt}
	\begin{table*}[!htbp]
	\centering
	\small
	\caption{Classification accuracies (\%) on \textbf{Office-Caltech}~(ResNet-101) and \textbf{Office-Home}~(ResNet-50) and \textbf{PACS} (ResNet-18) for \emph{multi-source UDA}. 
	}
	\label{table:multi}
	\vskip -0.05in
	\resizebox{0.92\textwidth}{!}{$
		\begin{tabular}{lccccaclccccaclcccca}
			\toprule
			(\textbf{Office-Caltech}) & $\to$A & $\to$C & $\to$D & $\to$W & Avg. & & (\textbf{Office-Home}) & $\to$Ar & $\to$Cl & $\to$Pr & $\to$Re & Avg. && (\textbf{PACS}) & $\to$A & $\to$C & $\to$P & $\to$S & Avg.\\
			\midrule
			ResNet-101 \cite{he2016deep} & 88.7 & 85.4 & 98.2 & 99.1 & 92.9 & & ResNet-50 \cite{he2016deep} & 65.3 & 49.6 & 79.7 & 75.4 & 67.5 & & ResNet-18 \cite{he2016deep} & 74.9 & 72.1 & 94.5 & 64.7 & 76.6 \\
			DAN \cite{long2015learning} & 91.6 & 89.2 & 99.1 & 99.5 & 94.8 & & M$^3$SDA-$\beta$ \cite{peng2019moment} & 67.2 & 58.6 & 79.1 & 81.2 & 71.5 && DANN~\cite{ganin2015unsupervised} & 81.9 & 77.5 & 91.8 & 74.6 & 81.5 \\
			DCTN \cite{xu2018deep} & 92.7 & 90.2 & 99.0 & 99.4 & 95.3 & & Meta-DANN~\cite{li2020online} & 70.6 & 59.1 & 80.2 & 82.8 & 73.2 && Meta-DANN~\cite{li2020online} & 87.3 & 84.9 & 96.9 & 73.2 & 85.6 \\
			MCD \cite{saito2018maximum} & 92.1 & 91.5 & 99.1 & 99.5 & 95.6 & & MCD~\cite{saito2018maximum} & 69.8 & 59.8 & 80.9 & 82.7 & 73.3 && Meta-MCD~\cite{li2020online} & 87.4 & 86.2 & 97.1 & 78.3 & 87.2 \\
			M$^3$SDA-$\beta$ \cite{peng2019moment} & 94.5 & 92.2 & 99.2 & 99.5 & 96.4 & & Meta-MCD~\cite{li2020online} & 70.2 & 60.5 & 81.2 & 83.4 & 73.8 && M$^3$SDA-$\beta$ \cite{peng2019moment} & 89.3 & 89.9 & 97.3 & 76.7 & 88.3 \\
			CMSS \cite{yang2020curriculum} & 96.0 & 93.7 & 99.3 & 99.6 & 97.2 & & SImpAl~\cite{venkat2020your} & 72.1 & \textbf{\color{blush}62.0} & 80.3 & 81.8 & 74.1 && CMSS~\cite{yang2020curriculum} & 88.6 & \textbf{\color{blush}90.4} & 96.9 & \textbf{\color{blush}82.0} & \textbf{\color{blush}89.5} \\
			\midrule
			Source-model-only & 95.4 & 93.6 & 98.9 & 98.4 & 96.6 & & Source-model-only & 67.3 & 51.2 & 78.7 & 81.4 & 69.6 && Source-model-only & 63.6 & 51.7 & 94.4 & 47.4 & 64.3 \\
			SHOT-IM & 96.3 & 95.5 & \textbf{\color{blush}99.6} & 99.8 & 97.8 & & SHOT-IM & 72.1 & 60.3 & 82.4 & 82.9 & 74.4 & & SHOT-IM & 89.6	& 87.9 & 98.6 & 62.5 & 84.7 \\
			SHOT & 96.2 & 96.2 & 98.5 & 99.8 & 97.7 & & SHOT & 73.0 & 60.4 & 83.9 & 83.3 & 75.2 & & SHOT & 90.7 & 88.1 & 98.5 & 75.4 & 88.2	\\
			\midrule
			Source-only++ & 96.3 & 95.5 & \textbf{\color{blush}99.6} & 99.8 & 97.8 & & Source-only++ & 70.3 & 51.6 & 83.0 & 83.5 & 72.1 & & Source-only++ & 72.1 & 44.1 & 98.1 & 41.5 & 63.9 \\
			SHOT-IM++ & \textbf{\color{blush}96.5} & \textbf{\color{blush}96.5} & 99.2 & 99.9 & \textbf{\color{blush}98.0} & & SHOT-IM++ & 72.3 & 60.4 & 82.5 & 83.2 & 74.6 & & SHOT-IM++ & 91.5 & 89.8 & \textbf{\color{blush}98.9} & 64.0 & 86.0 \\
			SHOT++ & 96.2 & \textbf{\color{blush}96.5} & 99.4 & \textbf{\color{blush}100.} & \textbf{\color{blush}98.0} & & SHOT++ & \textbf{\color{blush}73.1} & 61.3 & \textbf{\color{blush}84.3} & \textbf{\color{blush}84.0} & \textbf{\color{blush}75.7} & & SHOT++ & \textbf{\color{blush}92.3} & 89.7 & 98.8 & 75.5 & 89.1 \\ 
			\midrule
			Target-supervised & 96.7 & 95.3 & 99.0 & 98.9 & 97.5 & & Target-supervised & 74.5 & 77.9 & 91.4 & 84.4 & 82.0 & & Target-supervised & 92.7 & 92.9 & 98.4 & 93.8 & 94.5 \\  
			\bottomrule
		\end{tabular}
		$}
	\end{table*}

	\setlength{\tabcolsep}{2.0pt}
	\begin{table*}[htbp]
	\centering
	\small
	\caption{Classification accuracies (\%) on \textbf{Office-Home} and \textbf{VisDA-C} for \emph{partial-set UDA} (ResNet-50).}
	\label{table:pda}
	\vskip -0.05in
	\resizebox{0.92\textwidth}{!}{$
		\begin{tabular}{lccccccccccccaccca}
			\toprule
			Methods & \multicolumn{13}{c}{\textbf{Office-Home (65$\to$25})} & & \multicolumn{3}{c}{\textbf{VisDA-C (12$\to$6)}} \\
			\cmidrule{2-14}
			\cmidrule{16-18}
			Source$\to$Target & Ar$\to$Cl & Ar$\to$Pr & Ar$\to$Re & Cl$\to$Ar & Cl$\to$Pr & Cl$\to$Re & Pr$\to$Ar & Pr$\to$Cl & Pr$\to$Re & Re$\to$Ar & Re$\to$Cl & Re$\to$Pr & Avg. & & R$\to$S & S$\to$R & Avg. \\
			\midrule
			ResNet-50 \cite{he2016deep} & 46.3 & 67.5 & 75.9 & 59.1 & 59.9 & 62.7 & 58.2 & 41.8 & 74.9 & 67.4 & 48.2 & 74.2 & 61.3 & & 64.3 & 45.3 & 54.8 \\
			IWAN \cite{zhang2018importance} & 53.9 & 54.5 & 78.1 & 61.3 & 48.0 & 63.3 & 54.2 & 52.0 & 81.3 & 76.5 & 56.8 & 82.9 & 63.6 & & 71.3 & 48.6 & 60.0 \\
			SAN \cite{cao2018partiala} & 44.4 & 68.7 & 74.6 & 67.5 & 65.0 & 77.8 & 59.8 & 44.7 & 80.1 & 72.2 & 50.2 & 78.7 & 65.3 & & 69.7 & 49.9 & 59.8 \\
			DRCN~\cite{li2020deep_pda} & 54.0 & 76.4 & 83.0 & 62.1 & 64.5 & 71.0 & 70.8 & 49.8 & 80.5 & 77.5 & 59.1 & 79.9 & 69.0 & & 73.2 & 58.2 & 65.7 \\
			ETN \cite{cao2019learning} & 59.2 & 77.0 & 79.5 & 62.9 & 65.7 & 75.0 & 68.3 & 55.4 & 84.4 & 75.7 & 57.7 & 84.5 & 70.5 & & - & - & -\\
			SAFN \cite{xu2019larger} & 58.9 & 76.3 & 81.4 & 70.4 & 73.0 & 77.8 & 72.4 & 55.3 & 80.4 & 75.8 & 60.4 & 79.9 & 71.8 & & - & - & -\\
			RTNet$_{adv}$~\cite{chen2020selective} & 63.2 & 80.1 & 80.7 & 66.7 & 69.3 & 77.2 & 71.6 & 53.9 & 84.6 & 77.4 & 57.9 & 85.5 & 72.3 & & - & - & -\\
			BA$^3$US~\cite{liang2020balanced} & 60.6 & 83.2 & 88.4 & 71.8 & 72.8 & 83.4 & 75.5 & 61.6 & 86.5 & 79.3 & 62.8 & 86.1 & 76.0 & & - & - & -\\
			TSCDA~\cite{ren2020learning} & 63.6 & 82.5 & 89.6 & 73.7 & 73.9 & 81.4 & 75.4 & 61.6 & 87.9 & \textbf{\color{blush}83.6} & 67.2 & 88.8 & 77.4 & & - & - & -\\
			\midrule
			Source-model-only & 44.9 & 70.5 & 81.0 & 55.4 & 60.2 & 66.2 & 61.5 & 40.3 & 76.5 & 70.6 & 47.8 & 77.2 & 62.7 && 60.9 & 46.6 & 53.8 \\
			SHOT-IM & 59.1 & 83.9 & 88.5 & 72.7 & 73.5 & 78.4 & 75.9 & 59.9 & 90.3 & 81.3 & 68.6 & 88.7 & 76.7 && 69.2 & 68.8 & 69.0 \\
			SHOT & 64.6 & 85.1 & 92.9 & 78.4 & 76.8 & 86.9 & 79.0 & 65.7 & 89.0 & 81.1 & 67.7 & 86.4 & 79.5 && 73.1 & 74.2 & 73.6 \\
			\midrule
			Source-model-only++ & 50.3 & 77.1 & 86.6 & 66.2 & 67.6 & 75.7 & 69.2 & 46.4 & 83.6 & 76.2 & 51.3 & 82.4 & 69.4 && 67.7 & 65.8 & 66.8 \\
			SHOT-IM++ & 59.6 & 84.5 & 89.0 & 73.7 & 74.2 & 79.3 & 77.0 & 60.7 & \textbf{\color{blush}91.0} & 81.8 & \textbf{\color{blush}69.4} & \textbf{\color{blush}89.3} & 77.5 && 70.0 & 75.7 & 72.9 \\
			SHOT++ & \textbf{\color{blush}65.0} & \textbf{\color{blush}85.8} & \textbf{\color{blush}93.4} & \textbf{\color{blush}78.8} & \textbf{\color{blush}77.4} & \textbf{\color{blush}87.3} & \textbf{\color{blush}79.3} & \textbf{\color{blush}66.0} & 89.6 & 81.3 & 68.1 & 86.8 & \textbf{\color{blush}79.9} & & \textbf{\color{blush}75.3} & \textbf{\color{blush}78.6} & \textbf{\color{blush}77.0} \\
			\midrule
			Target-supervised & 81.0 & 91.5 & 85.8 & 80.0 & 91.5 & 85.8 & 80.0 & 81.0 & 85.8 & 80.0 & 81.0 & 91.5 & 84.6 && 98.8 & 89.9 & 94.3 \\
			\bottomrule
		\end{tabular}
		$}
	\end{table*} 

	As expected, on the medium-sized \textbf{Office-Home} dataset, our method SHOT++ significantly outperforms previously published state-of-the-art approaches, advancing the average accuracy from 70.4\% in GVB-GD~\cite{cui2020gradually} to 73.0\% in Table~\ref{table:home}. Besides, SHOT++ performs the best among 11 out of 12 separate tasks.
	For the transfer task Re$\to$Ar, SHOT++ gets the third-best result 73.7\% that is slightly lower than the best result 74.6\% of GVB-GD.
	Generally, the hypothesis transfer strategy works well enough, seen from the outperforming results of SHOT over prior methods,
	and the labeling transfer strategy further lifts the avg. accuracy by nearly 1 point.

	For the large-scale synthesis-to-real \textbf{VisDA-C} dataset, we follow the protocol in prior works \cite{saito2018adversarial,xu2019larger} and employ the most favoring backbone ResNet-101~\cite{he2016deep}. 
	As shown in Table~\ref{table:visda}, SHOT++ achieves the best per-class accuracy and wins among 8 out of 12 tasks.
	Even when ignoring the second stage, namely, labeling transfer, SHOT can still obtain a promising per-class result 85.5\%, higher than the prior state-of-the-art 82.7\% in STAR~\cite{lu2020stochastic}.
	Carefully comparing SHOT with prior work, we find that SHOT performs well even for the most challenging class `truck'.
	Besides, using the intra-domain semi-supervised learning stage via MixMatch, the per-class results are improved but the accuracy of the hard class `truck' decreases.
	This may be because large error in the labeled split affects the final results.

	\subsection{Results of Object Recognition beyond Vanilla UDA}
	\textbf{Results of object recognition for MSDA.}
	For the multi-source UDA setting, we adopt the protocol in \cite{yang2020curriculum} on \textbf{Office-Caltech} and \textbf{PACS} and the prototcol in \cite{li2020online} on \textbf{Office-Home}.
	For the three datasets, we specify a target subset and use other three subsets as three source domains, forming a multi-source UDA task.
	Likewise, SHOT does not access the source data but provided with multiple source models instead.
	The results of ours and previously published state-of-the-arts are shown in Table~\ref{table:multi}. 
	It is clear that SHOT achieves better results than CMSS~\cite{yang2020curriculum} or SImpAl~\cite{venkat2020your} in 3 out of 4 tasks on \textbf{Office-Caltech}, 3 of the 4 tasks on \textbf{PACS}, and 2 of the 4 tasks on \textbf{PACS}, respectively.
	With the incorporation of labeling transfer, SHOT++ wins SHOT for all these transfer tasks on the three datasets.
	Besides, the gap between SHOT and SHOT-IM is relatively small on \textbf{Office-Caltech} since the predictions learned by SHOT-IM are already good enough.
	On \textbf{PACS}, SHOT++ achieves competitive performance with that in CMSS~\cite{yang2020curriculum}.

	\setlength{\tabcolsep}{2.0pt}
	\begin{table*}[htbp]
	\centering
	\small
	\caption{Classification accuracies (\%) on \textbf{Office-Home} dataset for \emph{semi-supervised DA} (VGG16 on one-shot setting).}
	\label{tab:ssda}
	\vskip -0.05in
	\resizebox{0.9\textwidth}{!}{$
		\begin{tabular}{lcccccccccccca}
			\toprule
			SSDA (Source$\to$Target) & Ar$\to$Cl & Ar$\to$Pr & Ar$\to$Re & Cl$\to$Ar & Cl$\to$Pr & Cl$\to$Re & Pr$\to$Ar & Pr$\to$Cl & Pr$\to$Re & Re$\to$Ar & Re$\to$Cl & Re$\to$Pr & Avg. \\
			\midrule
			S+T~\cite{saito2019semi} & 37.5 & 63.6 & 69.5 & 51.4 & 65.9 & 64.5 & 52.0 & 37.0 & 71.6 & 61.2 & 39.5 & 75.3 & 57.4 \\
			DANN~\cite{ganin2015unsupervised} & 44.4 & 64.3 & 68.9 & 52.3 & 65.3 & 64.2 & 51.3 & 45.9 & 72.7 & 62.7 & 52.0 & 75.7 & 60.0 \\
			PAC~\cite{mishra2021surprisingly} & 43.5 & 69.8 & 69.5 & 45.3 & 69.6 & 65.3 & 55.3 & \textbf{\color{blush}54.7} & 73.1 & 64.6 & \textbf{\color{blush}56.4} & 78.8 & 62.2 \\
			MME~\cite{saito2019semi} & 45.8 & 68.6 & 72.2 & 57.5 & 71.3 & 68.0 & 56.0 & 46.2 & 74.4 & 65.1 & 49.1 & 78.7 & 62.7 \\
			ELP~\cite{huang2020effective} & 46.1 & 69.0 & 72.4 & 57.4 & 71.6 & 68.2 & 56.3 & 46.7 & 75.3 & 65.5 & 49.2 & 79.7 & 63.1 \\
			UODA~\cite{qin2021opposite} & 43.3 & 72.5 & 73.3 & 59.3 & 72.1 & 70.5 & 58.8 & 45.5 & 75.4 & \textbf{\color{blush}66.1} & 49.6 & 79.8 & 63.9\\
			\midrule
			labeled-data-only & 40.7 & 66.7 & 69.2 & 52.9 & 67.6 & 65.1 & 52.4 & 38.1 & 70.7 & 61.4 & 42.9 & 75.5 & 58.6 \\
			SHOT-IM & 47.5 & 72.4 & 74.1 & 59.4 & 73.3 & 71.2 & 57.9 & 45.2 & 76.5 & 64.5 & 49.6 & 80.6 & 64.4 \\
			SHOT & 49.1 & 73.9 & 74.9 & 59.4 & 75.0 & 72.9 & 58.0 & 47.0 & 77.1 & 65.0 & 50.7 & 80.8 & 65.3 \\
			\midrule
			labeled-data-only++ & 41.8 & 71.7 & 71.9 & 58.2 & 74.3 & 69.9 & 55.9 & 39.2 & 75.0 & 63.7 & 43.8 & 78.9 & 62.0 \\
			SHOT-IM++ & 48.1 & 73.6 & 75.3 & \textbf{\color{blush}60.5} & 74.6 & 72.1 & 58.9 & 45.6 & 76.7 & 64.8 & 50.2 & 81.4 & 65.2 \\
			SHOT++ & \textbf{\color{blush}49.7} & \textbf{\color{blush}75.0} & \textbf{\color{blush}76.0} & 60.4 & \textbf{\color{blush}76.1} & \textbf{\color{blush}73.6} & \textbf{\color{blush}59.8} & 47.5 & 
			\textbf{\color{blush}77.6} & 65.4 & 51.1 & 
			\textbf{\color{blush}81.7} & 
			\textbf{\color{blush}66.1} \\
			\midrule
			Target-supervised & 75.8 & 88.3 & 81.6 & 66.4 & 88.3 & 81.6 & 66.4 & 75.8 & 81.6 & 66.4 & 75.8 & 88.3 & 78.0 \\
			\bottomrule
		\end{tabular}
		$}
	\end{table*} 

	\textbf{Results of object recognition for PDA.}
	For the partial-set UDA setting, we follow the protocol in \cite{li2020deep_pda} on \textbf{Office-Home} and \textbf{VISDA-C}. 
	In particular, there are totally 25 classes (the first 25 in the alphabetical order) out of 65 classes in the target domain for \textbf{Office-Home}, while the first 6 classes in the alphabetical order out of 12 classes are included in the target domain for \textbf{VISDA-C}.
	Results of our methods and previous state-of-the-art PDA methods \cite{li2020deep_pda,chen2020selective,liang2020balanced,ren2020learning} are shown in Table~\ref{table:pda}.
	As explained in Section~\ref{sec:pda}, $\beta=0$ is utilized in all of our methods here.
	Compared with previous methods, SHOT obtains the best average accuracy for both datasets as before.
	Besides, SHOT again outperforms SHOT-IM by 2.8\% and 4.6\% in terms of the average accuracy on two datasets, and SHOT++ further improves the average accuracy from 79.5\% to 79.9\% and 73.6\% to 77.0\%, respectively.
	Generally, both the hypothesis transfer strategy and the labeling transfer strategy are proven effective for the challenging PDA problem.

	\textbf{Results of object recognition for SSDA.}
	For the semi-supervised domain adaptation setting, we follow the protocol in \cite{saito2019semi} on \textbf{Office-Home} under the one-shot setting where one labeled example per class is available in the target domain.
	As shown in Table~\ref{tab:ssda}, SHOT outperforms UODA~\cite{qin2021opposite} and MME~\cite{saito2019semi} in 10 out of 12 tasks and achieves the best average accuracy.
	Besides, SHOT is always superior to SHOT-IM, validating the effectiveness of self-supervision over the unlabeled target data.
	SHOT++ further improves the average accuracy from 65.3\% to 66.1\%, indicating the effectiveness of the labeling transfer strategy.
	
	\setlength{\tabcolsep}{3.0pt}
	\begin{table}[!tbp]
		\centering
		\caption{Accuracies for \textbf{ImageNet}$\to$\textbf{Caltech}. Methods $^\dagger$ utilize the training set of ImageNet besides pre-trained ResNet-50 model.}
		\label{tab:ic}
		\vskip -0.05in
		\resizebox{0.44\textwidth}{!}{$
			\begin{tabular}{ccccc}
				\toprule
				Methods$^\dagger$ & DRCN~\cite{li2020deep_pda} & SAN \cite{cao2018partiala} & IWAN \cite{zhang2018importance} & ETN \cite{cao2019learning} \\
				\midrule
				Accuracy & 75.3 & 77.8 & 78.1 & 83.2 $\pm$ 0.2 \\
				\midrule
				Methods & Source-only & SHOT-IM & SHOT ($\gamma_2=0$) & SHOT \\
				\midrule
				Accuracy & 69.7 & 81.8 $\pm$ 0.4 & 83.1 $\pm$ 0.1 & \textbf{\color{blush}83.3} $\pm$ 0.3 \\
				\bottomrule
			\end{tabular}
			$}
	\end{table}
	
	\textbf{Special case.} 
	One may wonder whether SHOT works if we cannot train the source model by ourselves.
	To find the answer, we utilize the most popular off-the-shelf pre-trained ImageNet models ResNet-50 \cite{he2016deep} and consider a special PDA task (\textbf{ImageNet}$\to$\textbf{Caltech}) to evaluate the effectiveness of SHOT with the same basic setting as \cite{cao2019learning}.
	Obviously, in Table \ref{tab:ic}, SHOT achieves a slightly higher mean accuracy than prior state-of-the-art ETN \cite{cao2019learning} even without access to the source data.
	It shows that the proposed hypothesis transfer strategy is indeed effective even without the design of model network architectures.

	\setlength{\tabcolsep}{2.0pt}
	\begin{table}[tbp]
	\centering
	\small
	\caption{Classification accuracies (\%) on large-scale \textbf{VisDA-C} dataset for \emph{vanilla closed-set UDA} (ResNet-50).}
	\label{table:visda-res50}
	\vskip -0.05in
	\resizebox{0.48\textwidth}{!}{$
		\begin{tabular}{lclc}
			\toprule
			Method & Per-class & Method & Per-class \\
			\midrule
			ResNet-50 \cite{he2016deep} & 52.4 & CDAN~\cite{long2018conditional} & 70.0 \\
			DANN~\cite{ganin2015unsupervised} & 57.4 & CDAN+TransNorm \cite{wang2019transferable} & 71.4 \\ 
			DAN~\cite{long2015learning} & 61.6 &  MDD~\cite{zhang2019bridging} & 74.6 \\ 
			MCD \cite{saito2018maximum} & 69.2 & GVB-GD~\cite{cui2020gradually} & 75.3 \\
			Dis-tune~\cite{liang2021distill} & 70.4 & DTA~\cite{lee2019drop} & 76.2 \\
			\midrule
			Source-model-only & 43.5 & Source-model-only++ & 52.2 \\
			SHOT-IM ($\beta=0$) & 64.6 & SHOT-IM++ ($\beta=0$) & 67.9 \\
			SHOT-IM & 73.9 & SHOT-IM++ & 75.6 \\
			SHOT ($\gamma_1=0$) & 74.6 & SHOT++ ($\gamma_1=0$) & 76.3 \\
			SHOT ($\gamma_2=0$) & 74.8 & SHOT++ ($\gamma_2=0$) & 76.5 \\
			SHOT & 76.7 & SHOT++ & \textbf{\color{blush}77.1} \\
			\midrule
			\multicolumn{3}{l}{Target-supervised (Synthesis $\to$ \textbf{Real})} & 88.8 \\
			\bottomrule
		\end{tabular}
		$}
	\end{table}
	
	\vspace{-7pt}
	\subsection{Model Analysis and Discussions}
	\textbf{Ablation study on different losses.} 	Following previous works \cite{zhang2019bridging,cui2020gradually}, we further adopt the ResNet-50~\cite{he2016deep} backbone to validate the effectiveness of our methods. Results are shown in Table~\ref{table:visda-res50}.
	With the hypothesis transfer strategy, SHOT beats the state-of-the-art method DTA~\cite{lee2019drop} by 0.5\% in terms of per-class accuracy.
	Benefited from the labeling transfer strategy, the per-class accuracy further grows from 76.7\% (SHOT) to 77.1\% (SHOT++) and again ranks the best for \textbf{VisDA-C} with the ResNet-50 backbone.
	
	In Table~\ref{table:visda-res50}, we further fix three balancing parameters (i.e., $\beta, \gamma_1, \gamma_2$) to zero in turn and investigate the effectiveness of each component within SHOT in Eq.~(\ref{eq:shot}), including $\mathcal{L}_{div}$, $\mathcal{L}_{ssl}^{1}$, and $\mathcal{L}_{ssl}^{2}$.
	Firstly, the advantages of SHOT-IM over SHOT-IM ($\beta=0$) validate the effectiveness of the diversity term $\mathcal{L}_{div}$. Incorporated with the labeling transfer strategy, SHOT-IM++ also obtains a better per-class accuracy than its variant SHOT-IM++ ($\beta=0$).
	Secondly, SHOT ($\gamma_1=0$) performs worse than SHOT, indicating the effectiveness of the self-supervised pseudo labeling term in Eq.~(\ref{eq:ssl1}).
	Thirdly, SHOT ($\gamma_2=0$) performs worse than SHOT, indicating the effectiveness of the self-supervised rotation prediction term in Eq.~(\ref{eq:ssl2}).
	Two latter conclusions can also be drawn by comparing SHOT ($\gamma_1=0$) and SHOT ($\gamma_2=0$) with SHOT-IM.
	Also, it seems $\mathcal{L}_{ssl}^{1}$ contributes more than $\mathcal{L}_{ssl}^{2}$ within SHOT.
	Finally, the benefits of the labeling transfer strategy are also easily validated by comparing the values in the second column with those in the fourth column.

	\begin{figure}[!htbp]
	\centering
	\footnotesize
	\setlength\tabcolsep{1mm}
	\includegraphics[width=0.85\linewidth,trim={0.3cm 0.5cm 0.3cm 0.3cm}, clip]{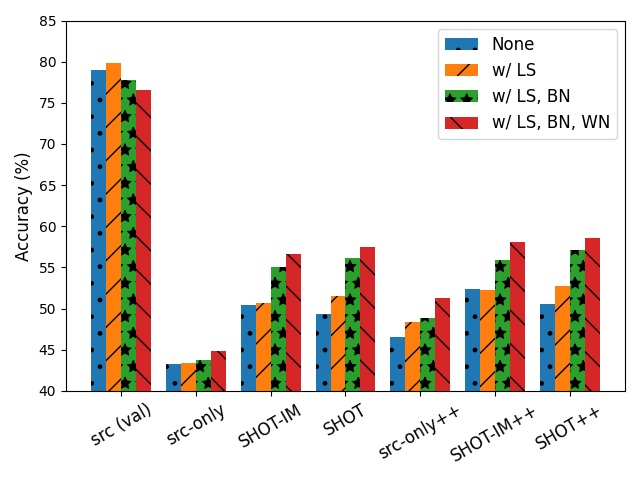}
	\caption{Ablation study of label smoothing (LS) and batch normalization (BN) and weight normalization (WN) in the network for a 65-way classification UDA task Ar$\to$Cl on \textbf{Office-Home}. `src (val)' denotes the accuracy in the \emph{source validation test}, and `src-only' is short for source-model-only. Best viewed in colors.}
	\label{fig:ab}
	\end{figure} 
	
	\textbf{Ablation study on network components.} As discussed in Section~\ref{sec:network}, we utilize label smoothing (LS), batch normalization (BN), and weight normalization (WN) during training the source model and learning the target feature encoder. 
	We report the ablation study about network components in Fig.~\ref{fig:ab} to validate their contribution.
	First, using BN or using WN results in the decreasing accuracy over the source training set, which may be useful for generalization since we find in the second bin, high accuracy on the source test set corresponds to low accuracy on the training set.
	Second, LS improves accuracies of both the source training set and the target set, which is desirable for source model generation.
	Third, the higher accuracy the `src-only' method obtains, the better results SHOT and its variants achieve. 
	Generally, BN and WN are beneficial to domain adaptation.
	The improvements brought by BN are larger than those brought by WN.
	
	\begin{figure}[!htb]
		\centering
		\footnotesize
		\setlength\tabcolsep{1mm}
		\renewcommand\arraystretch{0.1}
		\begin{tabular}{cc}
			\includegraphics[width=0.49\linewidth,trim={3.2cm 9.0cm 3.4cm 10.0cm}, clip]{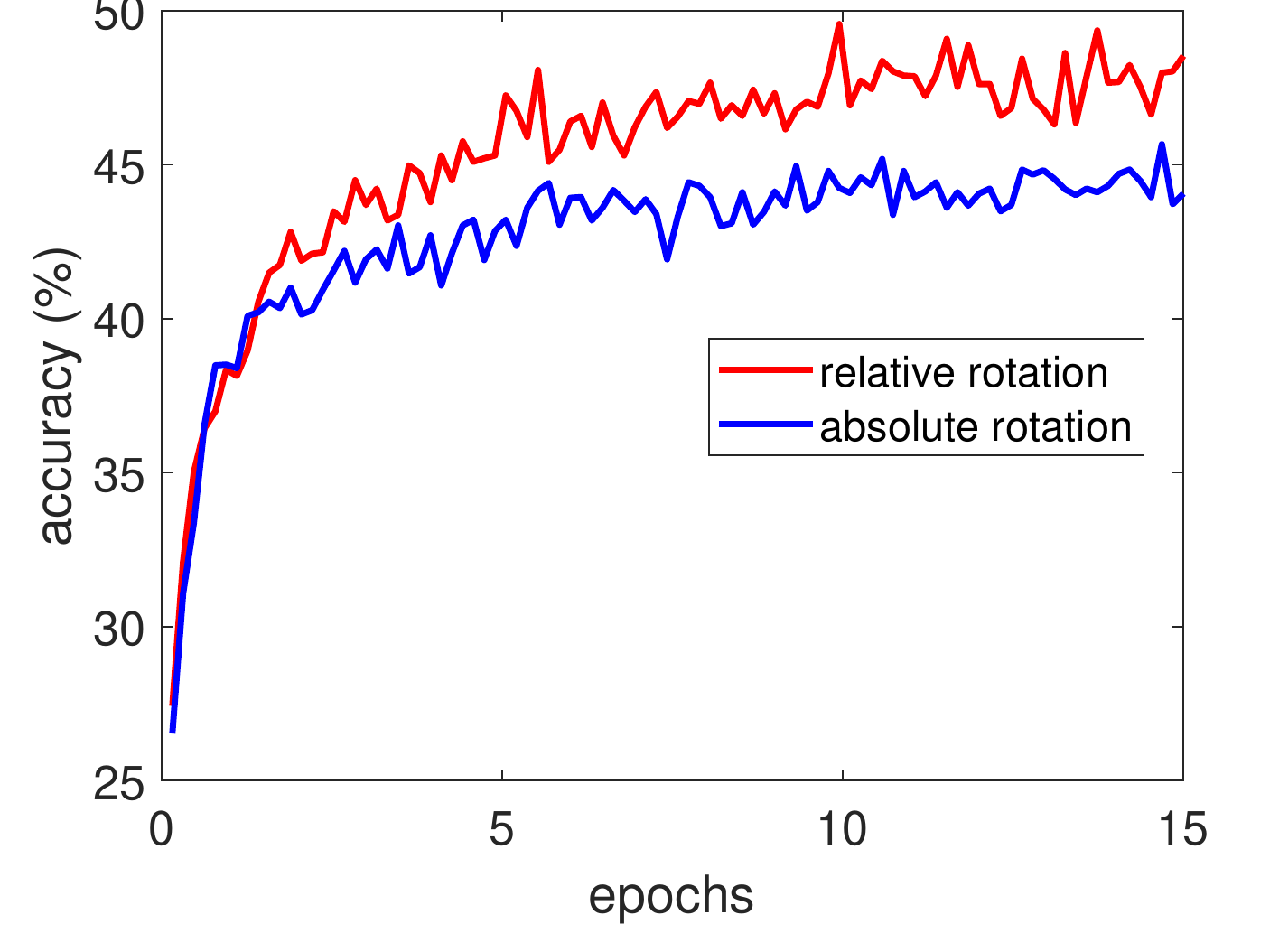} &
			\includegraphics[width=0.49\linewidth,trim={3.2cm 9.0cm 3.4cm 10.0cm}, clip]{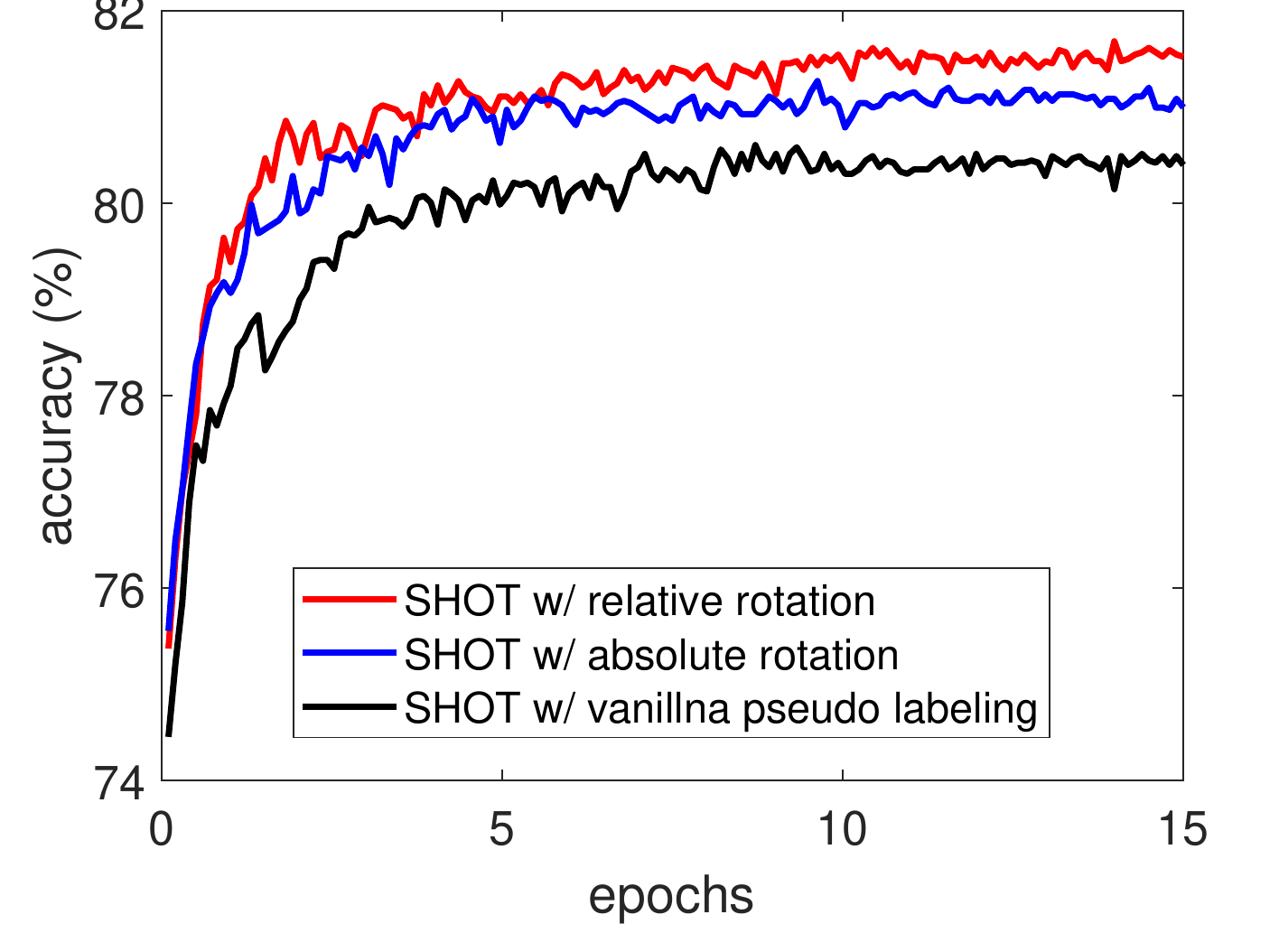} \\
			~\\
			(a) rotation prediction & (b) target accuracy
		\end{tabular}
		\caption{Accuracies of different variants during training for a 65-way classification UDA task Ar$\to$Re on \textbf{Office-Home} (15 epochs).}
		\label{fig:ap_ab}
	\end{figure} 
	
	\textbf{Discussion on loss functions.} 
	To analyze the advantages of our proposed self-supervised loss functions in Eq.~(\ref{eq:ssl1}) and Eq.~(\ref{eq:ssl2}), we design one vanilla alternative for each function on the transfer task Ar$\to$Re.
	As shown in Fig.~\ref{fig:ap_ab}(a), the proposed relative rotation prediction objective works better than the vanilla variant in terms of the rotation prediction accuracy. 
	Besides, comparisons in terms of the semantic accuracy in Fig.~\ref{fig:ap_ab}(b) indicate that the proposed relative rotation prediction objective is also beneficial to UDA.
	Compared with SHOT w/ vanilla pseudo-labeling, SHOT always obtains better results along the training process, implying the superiority of the proposed self-supervised pseudo-labeling term.

	\begin{figure}[!ht]
		\centering
		\footnotesize
		\setlength\tabcolsep{1mm}
		\renewcommand\arraystretch{0.1}
		\begin{tabular}{cc}
			\includegraphics[width=0.49\linewidth,trim={3.2cm 9.0cm 3.4cm 10.0cm}, clip]{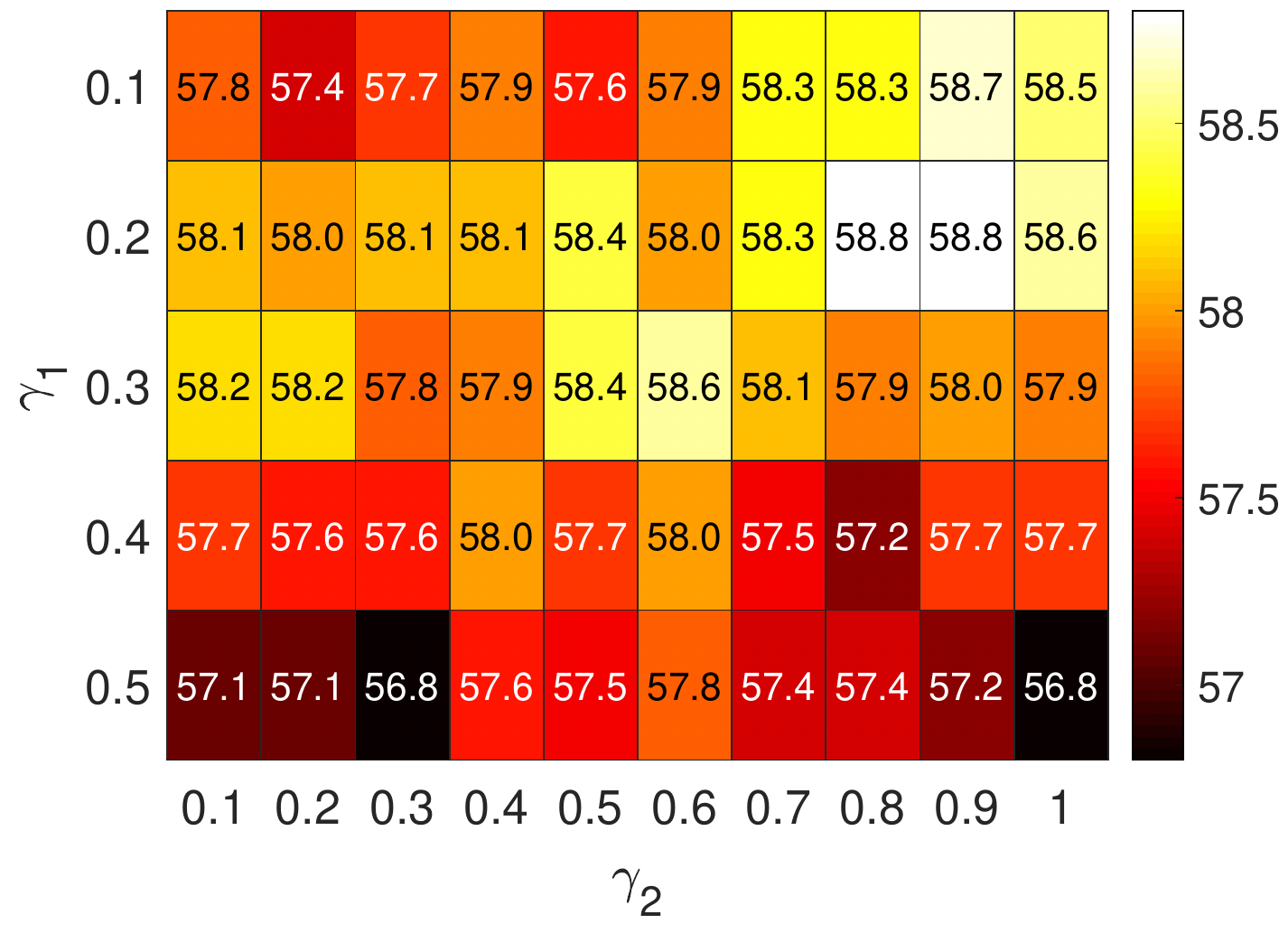} &
			\includegraphics[width=0.49\linewidth,trim={3.2cm 9.0cm 3.4cm 10.0cm}, clip]{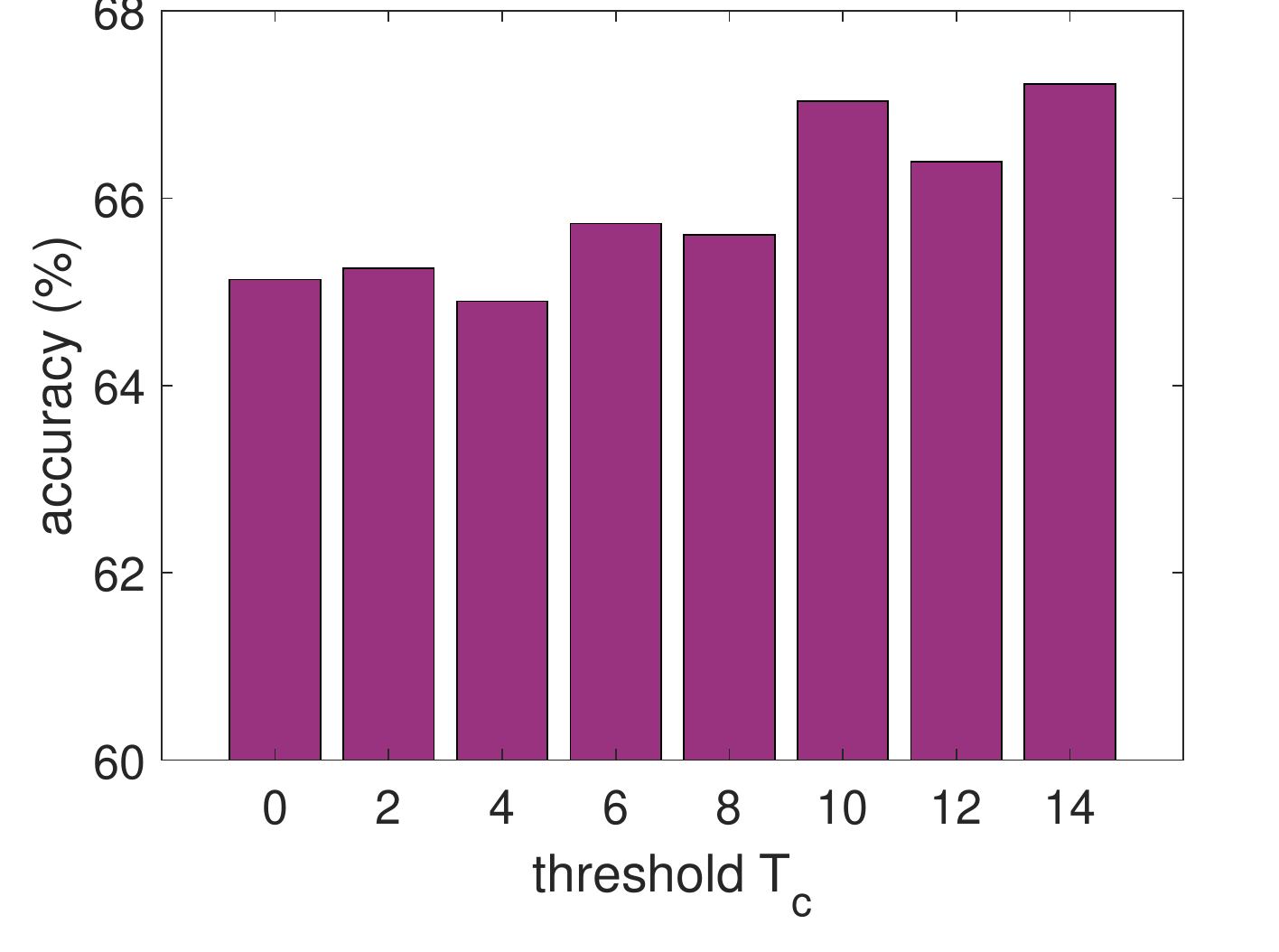} \\
			~\\
			(a) accuracy of Ar$\to$Cl (UDA) & (b) accuracy of Ar$\to$Cl (PDA)
		\end{tabular}
		\caption{Performance sensitivity of 3 parameters $\gamma_1$, $\gamma_2$, $T_c$ within SHOT.}
		\label{fig:parameter}
	\end{figure}
	
	\textbf{Parameter sensitivity.}
	To better understand the effects of $\gamma_1,\gamma_2$, we test their performance sensitivity in the UDA task Ar$\to$Cl on \textbf{Office-Home} and show the results in Fig.~\ref{fig:parameter}(a).
	The accuracies around $\gamma_1=0.2,\gamma_2=0.6$ are not sensitive.
	Besides, we study the sensitivity of the threshold parameter $T_c$ for the PDA task Ar$\to$Cl on \textbf{Office-Home} in Fig.~\ref{fig:parameter}(b). 
	It shows that the accuracies around $T_c=10$ are also not sensitive.
	Generally, the parameters within the proposed method i.e. SHOT are not sensitive.

	\begin{figure}[!ht]
	\centering
	\footnotesize
	\setlength\tabcolsep{1mm}
	\includegraphics[width=0.98\linewidth,trim={0.1cm 0.3cm 0.3cm 0.3cm}, clip]{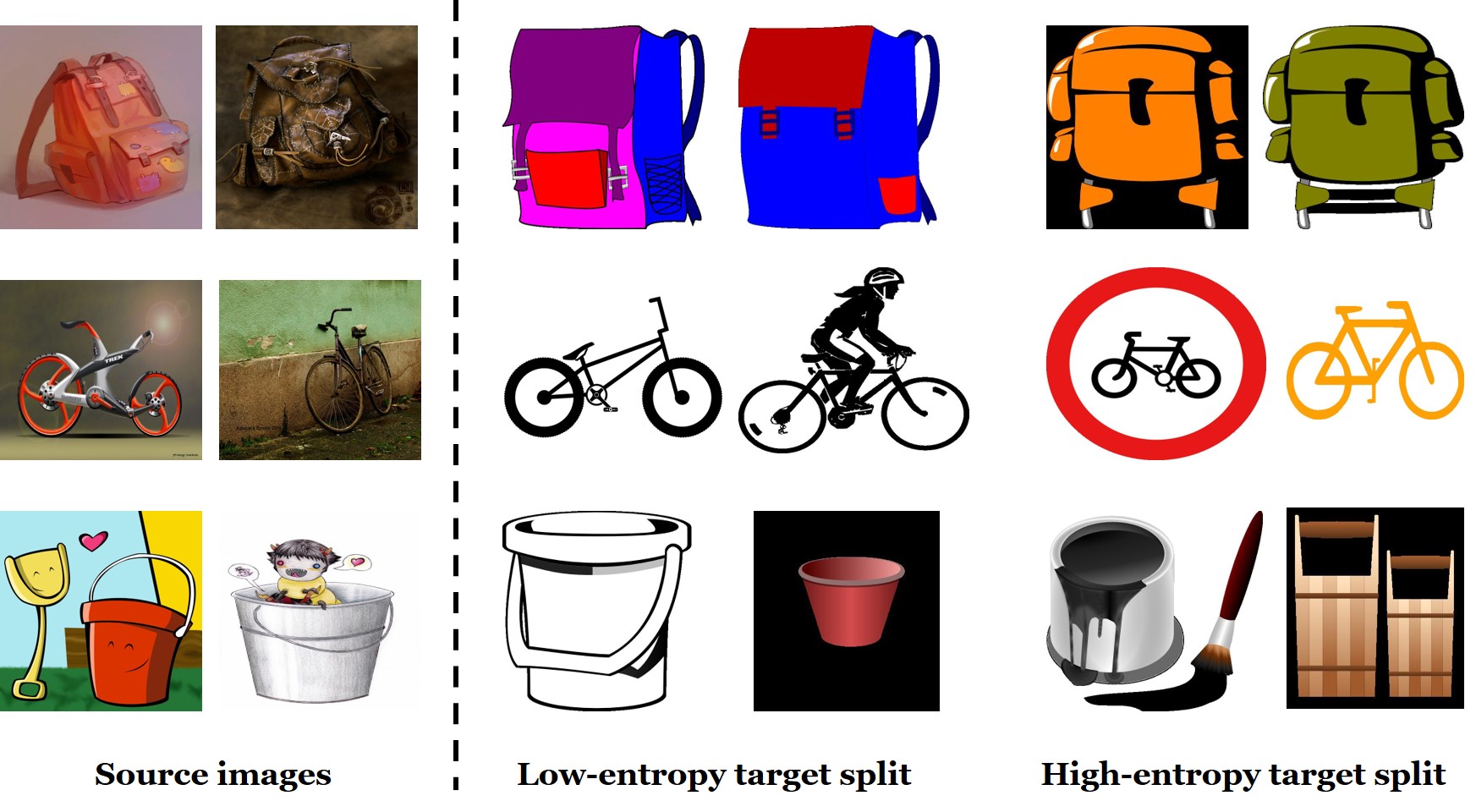}
	\caption{Images of the source domain, the low-entropy target split, and the high-entropy target split for Ar$\to$Cl on \textbf{Office-Home} (closed-set).}
	\label{fig:target_split}
	\end{figure} 

	\textbf{Qualitative Study.} We randomly select some samples in the source domain, the low-entropy target split, and the high-entropy target split to provide some intuitive insights about the labeling transfer strategy.
	Particularly, we pick up two images from three representative classes, i.e., `backpack', `bike', and `bucket', for the UDA task Ar$\to$Cl on \textbf{Office-Home}, and show them in Fig.~\ref{fig:target_split}.
	It can be seen that the proposed strategy can well separate the easy samples from the hard samples in the target domain.
	Besides, the easy samples in the low-entropy target split are more trustworthy than the source samples for the hard samples in the high-entropy target split, making the proposed labeling transfer strategy understandable and effective.

	\begin{figure*}[!htbp]
	\centering
	\footnotesize
	\setlength\tabcolsep{1mm}
	\renewcommand\arraystretch{0.1}
	\begin{tabular}{cccc}
		\includegraphics[width=0.24\linewidth,trim={3.2cm 9.0cm 4.2cm 10.0cm}, clip]{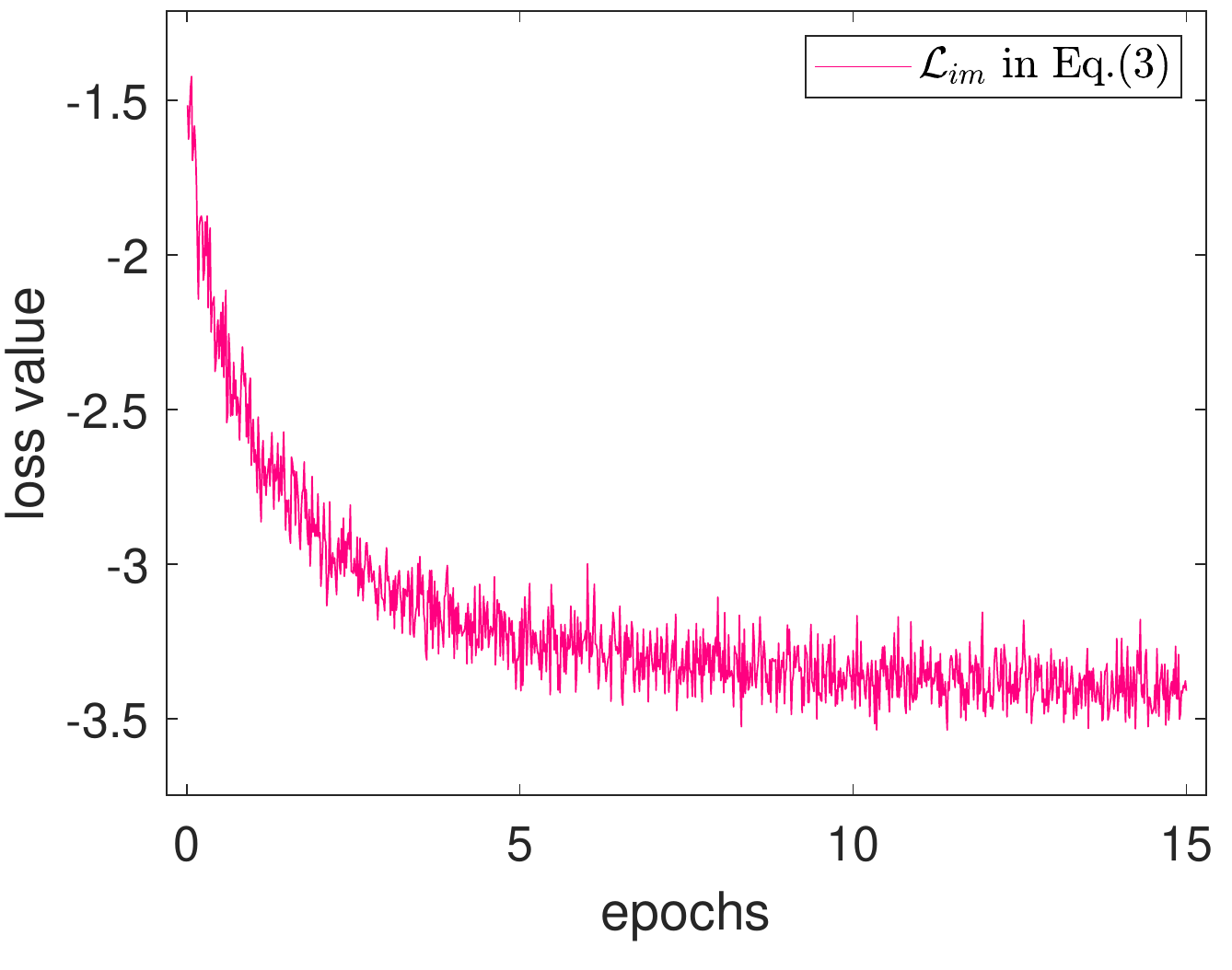} &
		\includegraphics[width=0.24\linewidth,trim={3.2cm 9.0cm 4.2cm 10.0cm}, clip]{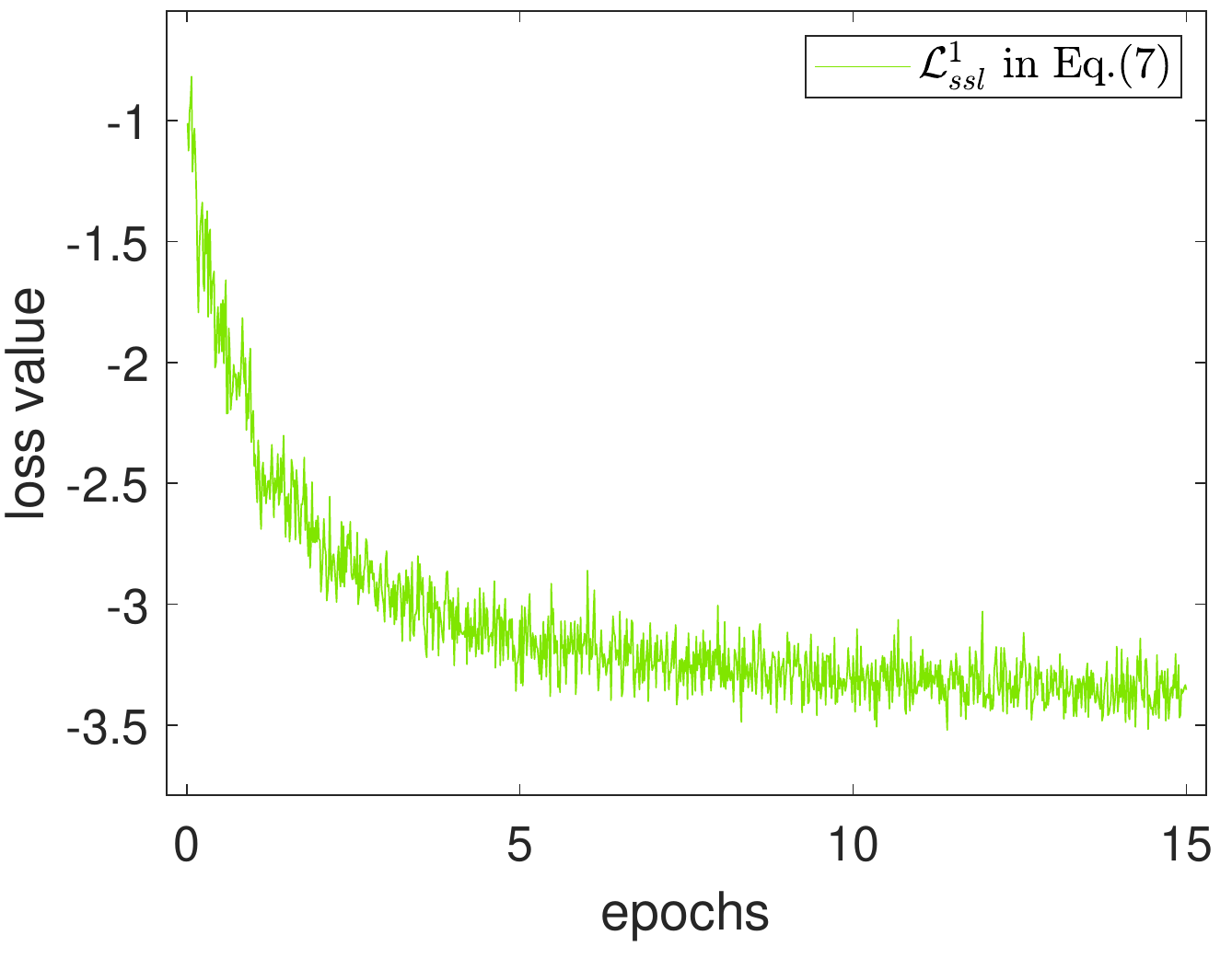} &
		\includegraphics[width=0.24\linewidth,trim={3.2cm 9.0cm 4.2cm 10.0cm}, clip]{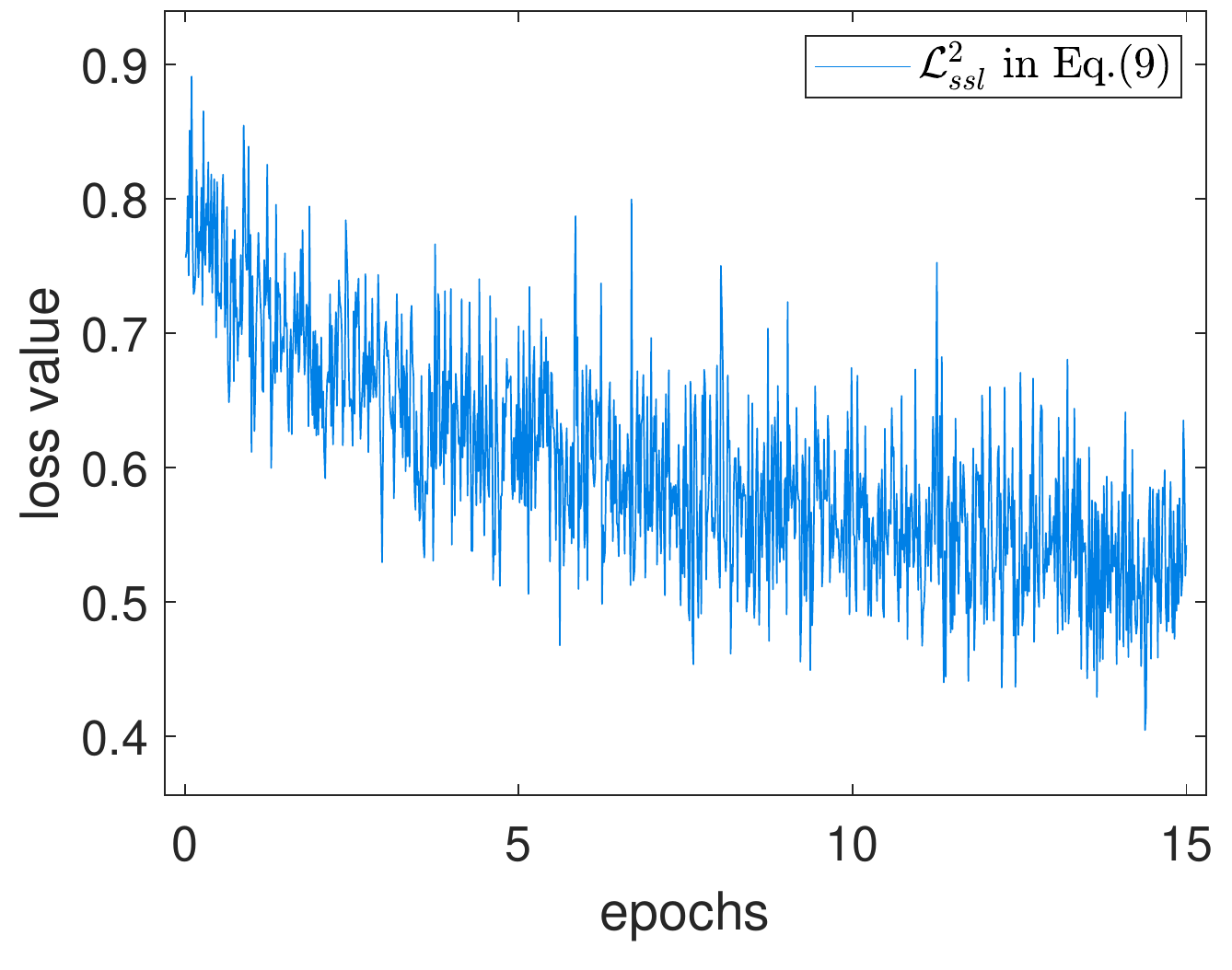} &
		\includegraphics[width=0.24\linewidth,trim={3.2cm 9.0cm 4.2cm 10.0cm}, clip]{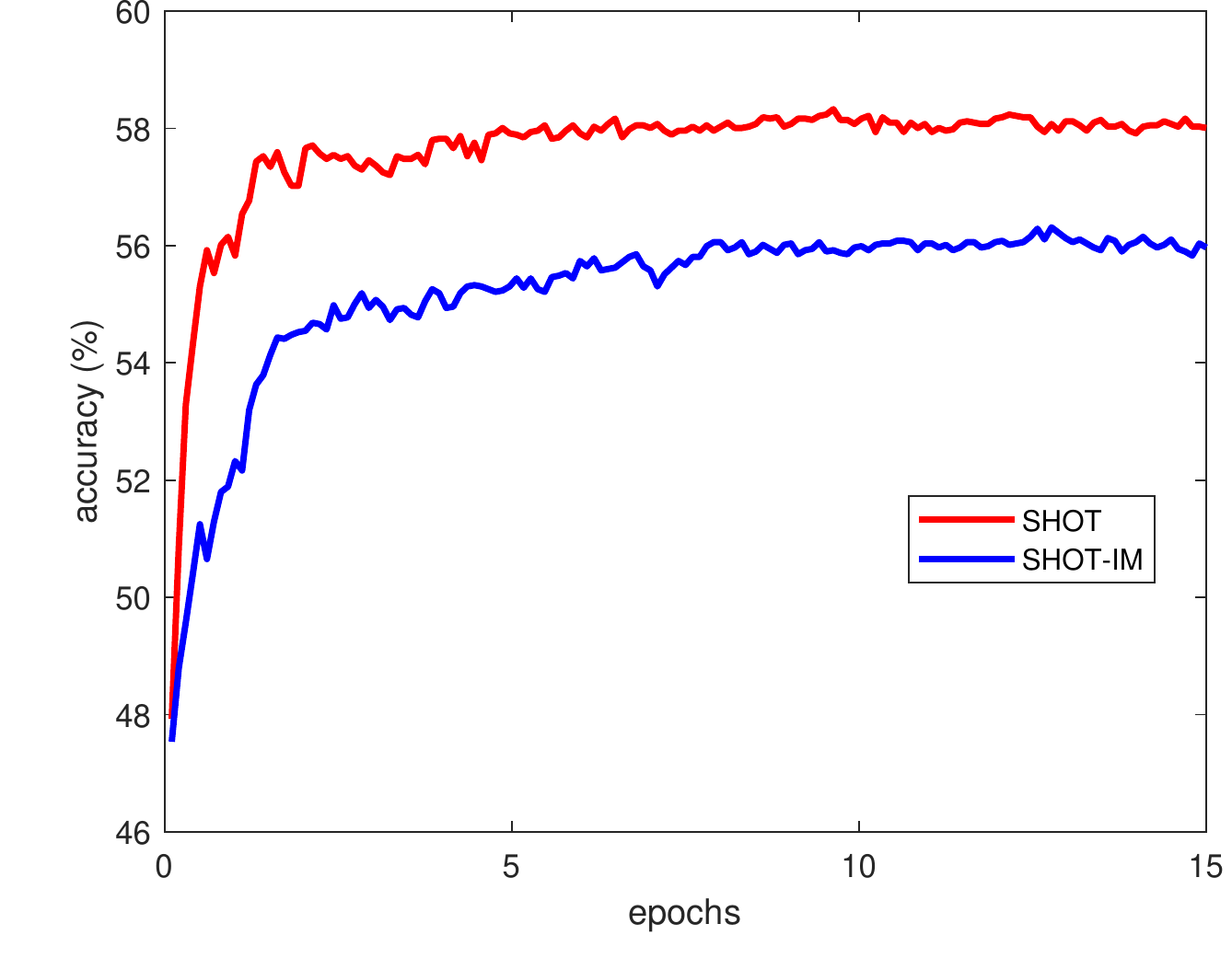} \\
		~\\
		(a) value of $\mathcal{L}_{im}$ in Eq.~(\ref{eq:im}) & (b) value of $\mathcal{L}_{ssl}^{1}$ in Eq.~(\ref{eq:ssl1})& (c) value of $\mathcal{L}_{ssl}^{2}$ in Eq.~(\ref{eq:ssl2}) & (d) Accuracy (\%)
	\end{tabular}
	\caption{Values of different loss functions and the accuracy during training for a 65-way classification UDA task Ar$\to$Cl on \textbf{Office-Home} (15 epochs).}
	\label{fig:conv_ac}
	\end{figure*} 

	\begin{figure*}[!htbp]
		\centering
		\footnotesize
		\setlength\tabcolsep{1mm}
		\renewcommand\arraystretch{0.1}
		\begin{tabular}{ccc}
			\includegraphics[width=0.32\linewidth,trim={3.2cm 10.0cm 3.4cm 10.0cm}, clip]{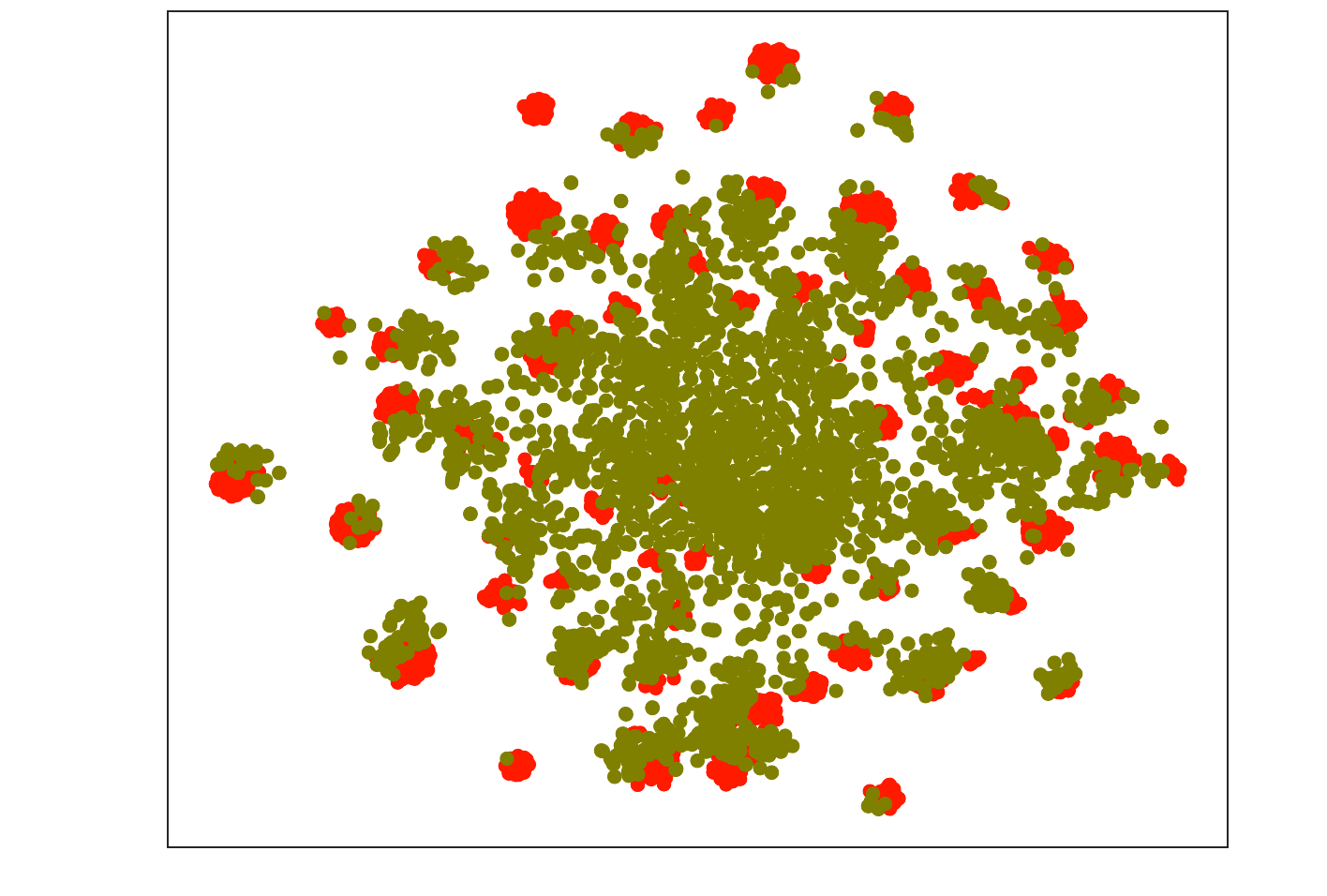} &
			\includegraphics[width=0.32\linewidth,trim={3.2cm 10.0cm 3.4cm 10.0cm}, clip]{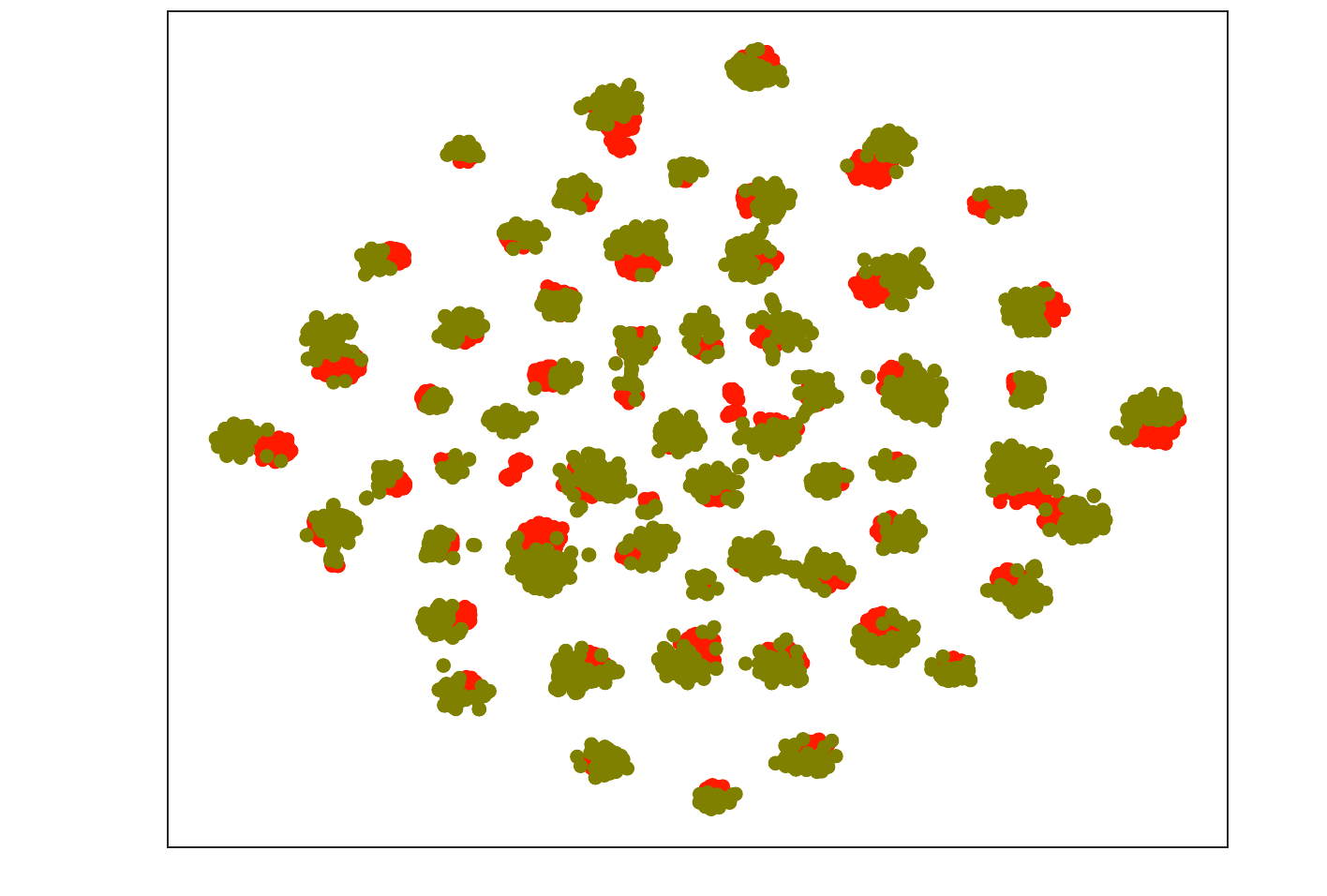} &
			\includegraphics[width=0.32\linewidth,trim={3.2cm 10.0cm 3.4cm 10.0cm}, clip]{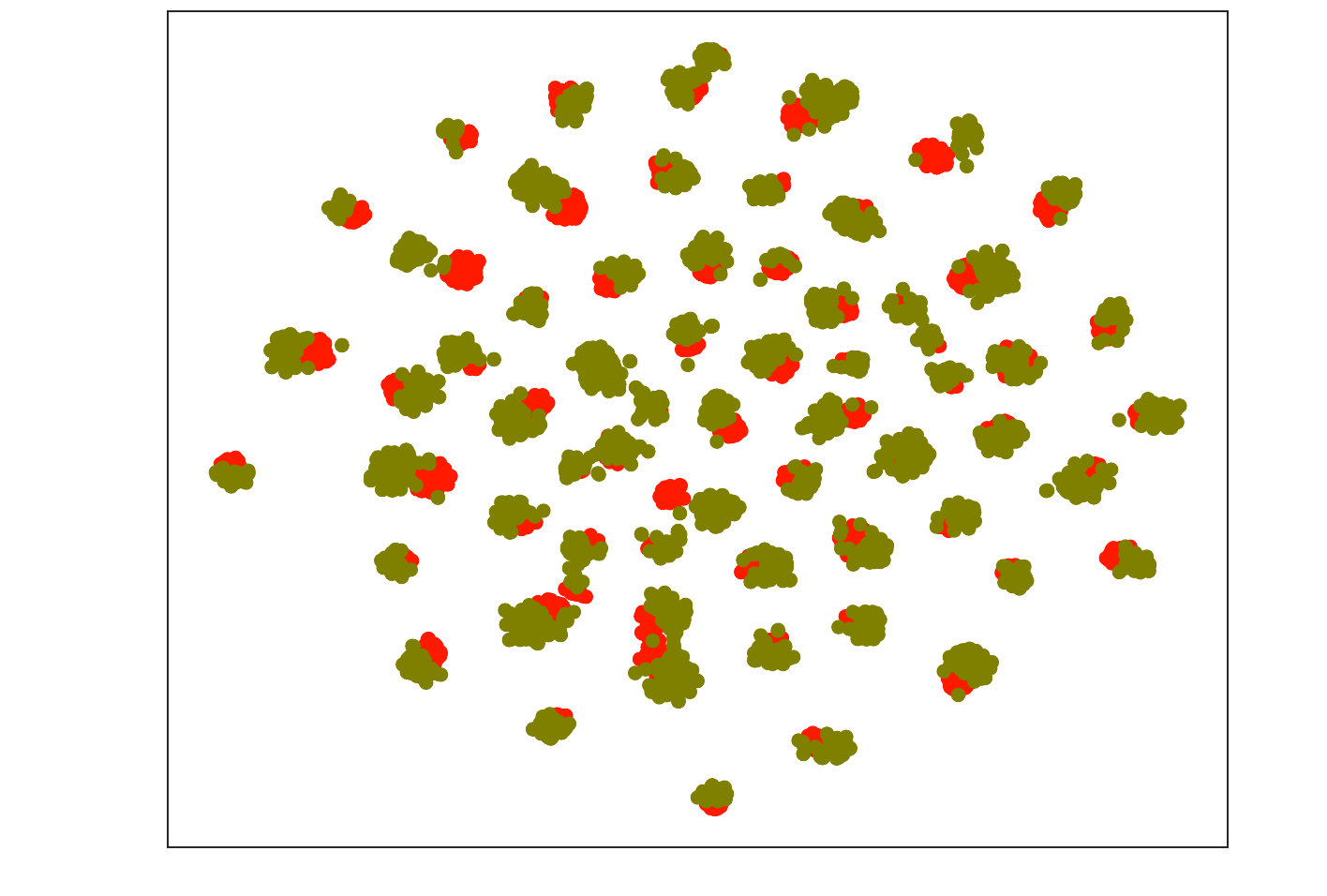} \\
			~\\
			(a) Source-model-only & (b) SHOT-IM & (c) SHOT
		\end{tabular}
		\caption{The t-SNE feature visualizations for a 65-way classification UDA task Ar$\to$Cl on \textbf{Office-Home}. \text{\color{red}{Circles in red}} denote unseen source data and \text{\color{olive}{circles in olive}} denote target data. Best viewed in colors.}
		\label{fig:tsne_AC}
	\end{figure*} 
	
	\begin{figure*}[!htbp]
		\centering
		\footnotesize
		\setlength\tabcolsep{1mm}
		\renewcommand\arraystretch{0.1}
		\begin{tabular}{ccc}
			\includegraphics[width=0.32\linewidth,trim={3.2cm 10.0cm 3.4cm 10.0cm}, clip]{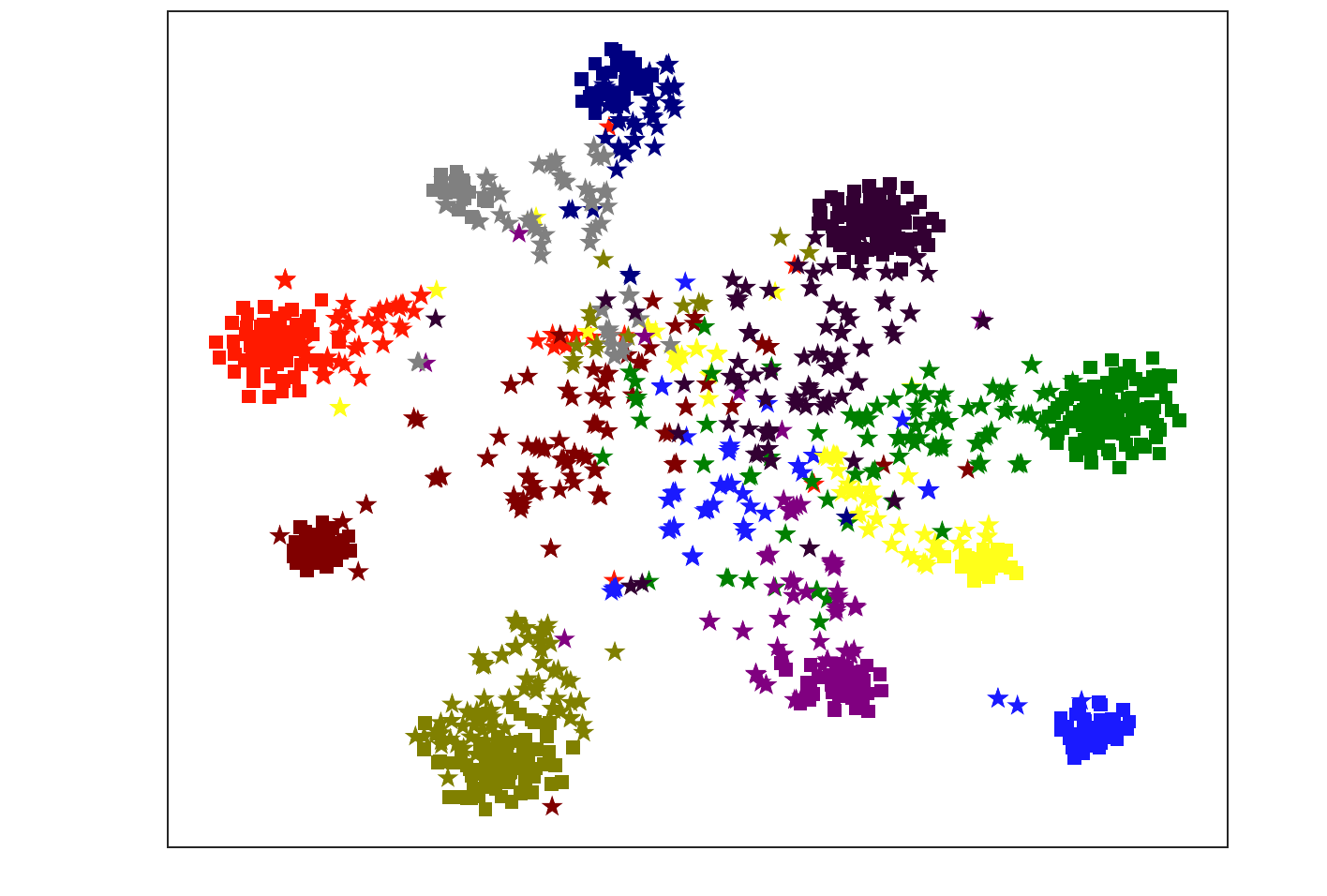} &
			\includegraphics[width=0.32\linewidth,trim={3.2cm 10.0cm 3.4cm 10.0cm}, clip]{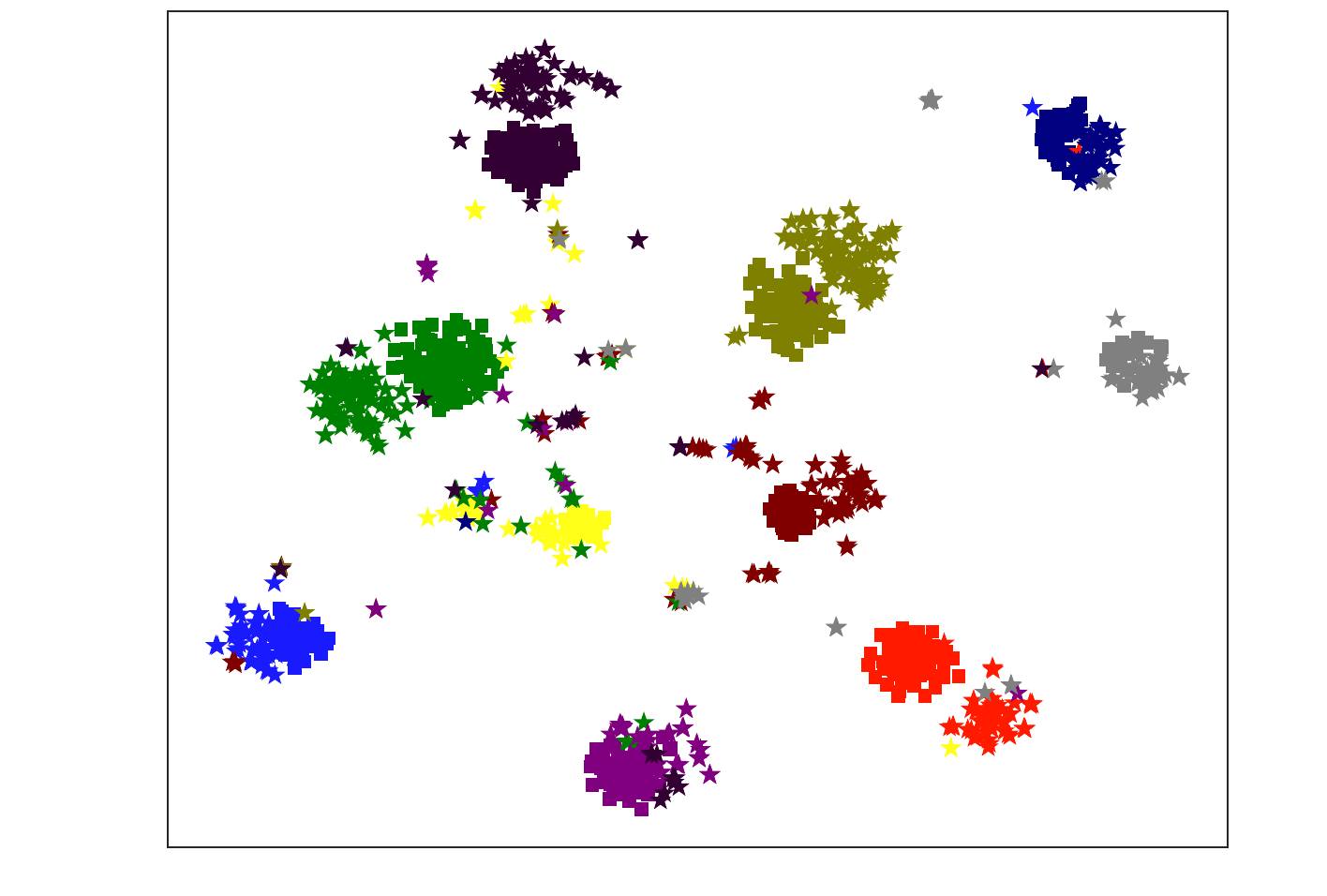} &
			\includegraphics[width=0.32\linewidth,trim={3.2cm 10.0cm 3.4cm 10.0cm}, clip]{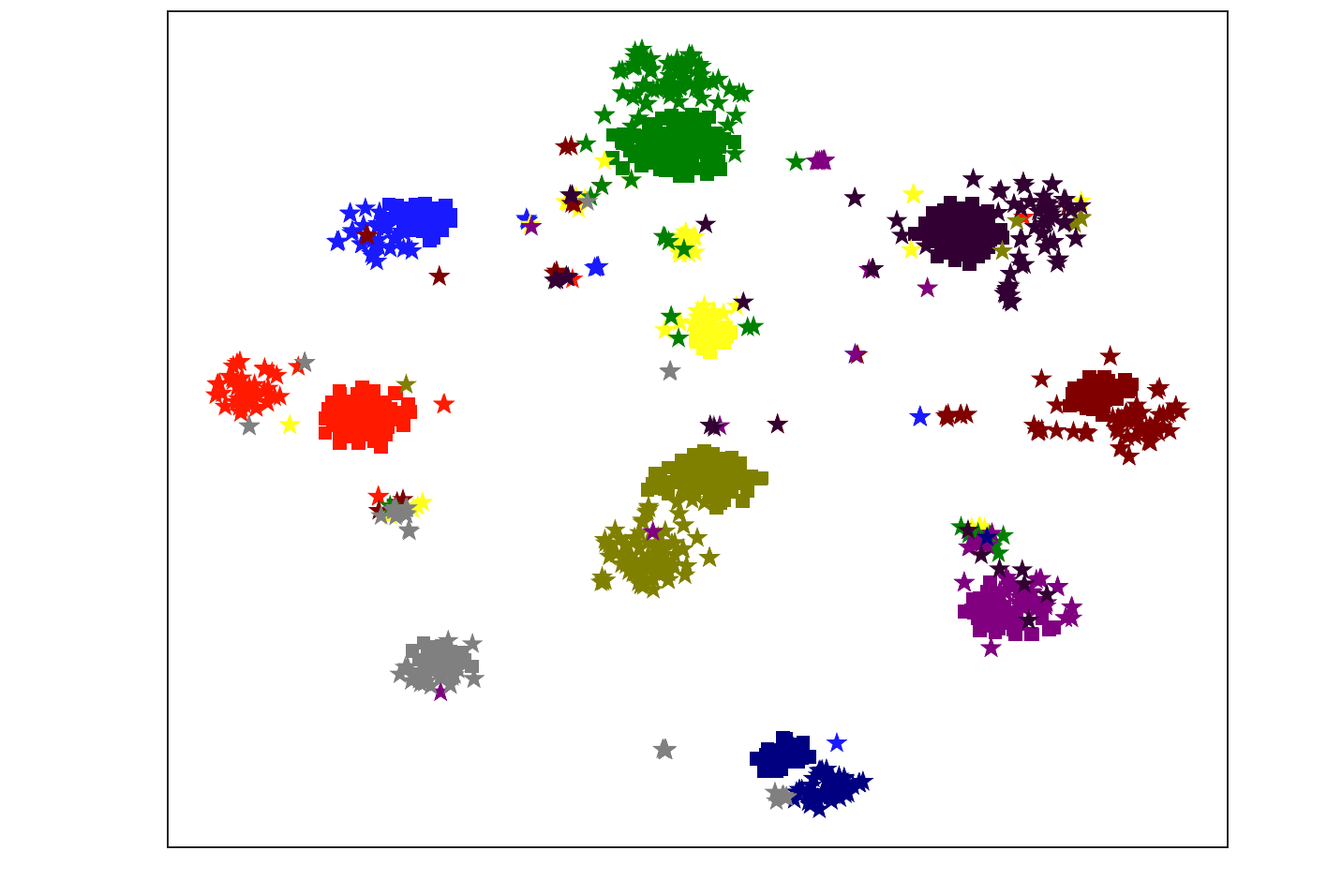} \\
			~\\
			(a) Source-model-only & (b) SHOT-IM & (c) SHOT
		\end{tabular}
		\caption{The t-SNE feature visualizations for a 65-way classification UDA task Ar$\to$Cl on \textbf{Office-Home}. For a better illustration, we choose features in the \textbf{first 10 classes} of each domain, and different color denotes different class. Best viewed in colors. [source in square, target in star]}
		\label{fig:tsne_AC10}
	\end{figure*} 

	\textbf{Training stability.} We investigate the accuracy and the values of three different objective functions within the optimization process in Fig.~\ref{fig:conv_ac} on the UDA task Ar$\to$Cl.
	It can be easily seen that the values of $\mathcal{L}_{im}$ and $\mathcal{L}_{ssl}^{1}$ quickly decrease and converge after nearly 8 epochs. 
	The value of rotation prediction loss $\mathcal{L}_{ssl}^{2}$ also keeps decreasing but at a slow speed.
	As shown in Fig.~\ref{fig:conv_ac}(d), the accuracy varies following a very similar trend, i.e. growing up quickly and starting to converge after 6 epochs.
	Generally, the training procedure of SHOT is stable and effective.

	\textbf{Feature visualization.}
	We provide the t-SNE visualizations \footnote{\url{https://lvdmaaten.github.io/tsne/}} of the features learned by Source-model-only, SHOT-IM, and SHOT for the UDA task Ar$\to$Cl on \textbf{Office-Home} in Fig.~\ref{fig:tsne_AC} and Fig.~\ref{fig:tsne_AC10}, respectively.
	As expected, both SHOT-IM and SHOT help align the target features with the source features in Fig.~\ref{fig:tsne_AC}.
	Carefully looking at the semantic labels in Fig.~\ref{fig:tsne_AC10}, we find that SHOT outperforms SHOT-IM by semantically aligning features from different domains.
	
	\section{Conclusion}
	In this paper, we have proposed a generic representation learning framework called source hypothesis transfer (SHOT) for source data-absent unsupervised domain adaptation.
	SHOT merely needs the well-trained source model and offers the feasibility of unsupervised domain adaptation without access to the source data which may be private or decentralized.
	Specifically, SHOT learns the optimal target-specific feature learning module to fit the source hypothesis by exploiting information maximization and self-supervised learning.
	We further present a labeling transfer strategy and apply it to enhance SHOT to SHOT++, which exploits the intra-domain information via a semi-supervised algorithm.
	Experiments for both digit classification and object recognition verify that SHOT and SHOT++ can achieve results comparable to or even better than the state-of-the-art for three different unsupervised domain adaptation scenarios as well as the semi-supervised domain adaptation problem.
	In the future, we plan to apply the proposed methods to other visual tasks like semantic segmentation \cite{zhang2019curriculum} and object detection \cite{chen2018domain}.

	\ifCLASSOPTIONcompsoc
	 \section*{Acknowledgments}
	\else
	 \section*{Acknowledgment}
	\fi
	The authors would like to thank the reviewers and the associate editor for their valuable comments. 
	The authors also thank Quanhong Fu and Weihao Yu for their help to improve the technical writing aspect of this paper. This work is supported by AISG-100E-2019-035, MOE2017-T2-2-151 and CRP20-2017-0006.
	
	 \ifCLASSOPTIONcaptionsoff
	 \newpage
	 \fi
	
	\bibliographystyle{IEEEtran}
	\bibliography{example_paper}
	
	\vfill
\end{document}